\documentclass{article}

\PassOptionsToPackage{numbers, compress}{natbib}

\usepackage[dandb, final]{neurips_2025}
\usepackage{soul}

\usepackage[utf8]{inputenc} 
\usepackage[T1]{fontenc}    
\usepackage{hyperref}       
\usepackage{url}            
\usepackage{booktabs}       
\usepackage{amsfonts}       
\usepackage{nicefrac}       
\usepackage{microtype}      
\usepackage{xcolor}         
\usepackage[inline]{enumitem}
\usepackage{amsmath}
\usepackage{minted}
\usepackage{fvextra}
\usepackage{standalone}
\usepackage{mystyle}
\usepackage{multirow}
\usepackage{wrapfig}
\usepackage{subcaption}
\usepackage{tikz}

\usepackage{caption}
\usepackage{makecell}
\usepackage{float}

\usepackage{xr-hyper}

\usetikzlibrary{
  positioning,
  calc,
  decorations.pathreplacing
}

\title{Quantifying Generalisation in Imitation Learning}

\author{%
  Nathan Gavenski\\
  Department of Informatics \\
  King's College London \\
  \texttt{nathan.schneider\_gavenski@kcl.ac.uk} \\
  \And
  Odinaldo Rodrigues \\
  Department of Informatics \\
  King's College London \\
  \texttt{odinaldo.rodrigues@kcl.ac.uk} \\
}

\begin{document}

\maketitle

\begin{abstract}
Imitation learning benchmarks often lack sufficient variation between training and evaluation, limiting meaningful generalisation assessment. 
We introduce Labyrinth, a benchmarking environment designed to test generalisation with precise control over structure, start and goal positions, and task complexity.
It enables verifiably distinct training, evaluation, and test settings.
Labyrinth provides a discrete, fully observable state space and known optimal actions, supporting interpretability and fine-grained evaluation.
Its flexible setup allows targeted testing of generalisation factors and includes variants like partial observability, key-and-door tasks, and ice-floor hazards.
By enabling controlled, reproducible experiments, Labyrinth advances the evaluation of generalisation in imitation learning and provides a valuable tool for developing more robust agents.
\end{abstract}

\section{Introduction} \label{sec:intro}

Imitation learning lies at the intersection of reinforcement and supervised learning.
It can be seen as a relaxation of the reinforcement learning problem, where the agent learns a new skill through its own experiences, into a supervised learning setting, where the agent learns by observing others perform the same task.
Like supervised learning, imitation learning relies on observed data for training.
However, its agentic nature makes evaluation more akin to reinforcement learning, as the agent's performance is assessed through interactions with an environment rather than static comparisons against a dataset.
As a result, many common evaluation benchmarks for imitation learning originate from the field of reinforcement learning.
The most common benchmarks~\cite{gavenski2024imitation} for imitation learning are:
\begin{enumerate*}[label=(\roman*)]
    \item CartPole~\cite{barto1983neuronlike}; and 
    \item MountainCar~\cite{moore1990efficient}, which are classic control tasks;
    \item Ant~\cite{schulman2015high}; and
    \item Humanoid~\cite{tassa2018deepmind}, which are continuous control tasks; and
    \item Atari Games, which set a benchmark for various games.
\end{enumerate*}

Classic control tasks, although reasonable for testing the initial capabilities of an imitation learning agent, are too simplistic to capture the complexities of real-world decision-making.
They have low-dimensional state spaces and a limited range of discrete actions.
As a result, they provide only a narrow evaluation of an imitation learning agent's capabilities.
On the other hand, Continuous control tasks provide a more challenging evaluation setting.
These environments feature high-dimensional state and action spaces, requiring agents to learn complex motor control strategies.
However, they still share key limitations with classic control tasks, such as a lack of precise state abstractions and the expected behaviour for the agent at any given state.
The first limitation refers to the vector state lacking information about the environment setting, such as the length of limbs for robots and the goal position, since the assumption is that the learned agent will be evaluated under the exact same constraints as it was during training.
A common solution to incorporate this information is to use image-based states.
However, when using images as states, the state may not accurately represent the difference between states due to the loss of precision from continuous numbers to pixel-based representation, and may exhibit partial observability since some parts of the agent may not be visible for the entire time.
For the second limitation, in these environments, finding the optimal expected behaviour for an agent in any given state is virtually impossible, which hinders the formal assessment of generalisation.
Finally, Atari Games introduces diverse tasks with visual inputs and long-term strategic planning.
While they provide a more varied and challenging benchmark, they remain constrained because the training data and test environments do not differ, meaning agents are evaluated under the same conditions in which they were trained.
This prevents a clear separation between training and testing data, which is crucial for assessing generalisation.

To address these limitations, we introduce Labyrinth\footnote{Source code available at: \url{https://github.com/NathanGavenski/Labyrinth}}, a novel environment designed to:
\begin{enumerate*}[label=(\roman*)]
    \item explicitly separate training and test data by altering structure, goals, or starting positions, demanding generalisation;
    \item provide a discrete and fully observable state space, where all possible states, transitions, and optimal actions are explicitly defined, enabling precise analysis of an agent's decision-making;
    \item allow for the systematic analysis of an agent's ability to learn and adapt to structural changes, offering insights into its robustness and generalisation capabilities; and 
    \item the environment can be easily customised to increase the difficulty further, e.g., by increasing the size of the labyrinth or maintaining the same solution set but changing the structure to analyse the inner parameters of the agent.
\end{enumerate*}
Labyrinth offers a more robust and comprehensive benchmark for imitation learning, more effectively capturing the challenges of real-world learning scenarios that require drastic adaptation from the agent than existing environments.

\section{Labyrinth Environment} \label{sec:environment}

In this work, we propose the Labyrinth environment to help assess the generalisation capabilities of imitation learning agents.
Navigating through a labyrinth from designated starting and goal positions by observing the labyrinth's entire structure is a trivial task. 
Humans can find a route by analysing all paths connecting the start and goal positions, and then applying a given criterion to select one (e.g., the shortest).
Classical problem-solving approaches, such as breadth-first search, can generalise to any labyrinth structure (considering the problem's solution and not its optimality). 
Therefore, navigating through a labyrinth should be considered an easy and well-suited task for measuring how well an imitation learning algorithm learns, and how general the agent's resulting capability is, i.e., by using structural configurations (e.g., wall locations, obstacles, etc) not present in the training data or moving from a different initial starting point.

Unlike other environments, a labyrinth offers some inherent characteristics:
\begin{enumerate*}[label=(\roman*)]
    \item agents cannot perform state-matching by forcing a path to be similar to its training data;
    \item changing the configuration of the labyrinth (walls, start and goal) does not affect the task and is easy to define; and
    \item changes between states are easier to identify since states can only differ by the agent's position.
\end{enumerate*}
These characteristics allow us to perform a more systematic evaluation of different methods.
For example, one can train a set of agents in one labyrinth structure and only the starting position or a subset of its walls.
Alongside the traditional task of navigating a labyrinth to reach a goal, the Labyrinth environment offers the possibility of making solutions more complex by adding two additional components: 
\begin{enumerate*}[label=(\roman*)]
    \item \textit{key and door}, where the agent must retrieve a key to open a door before reaching the goal; and 
    \item \textit{ice floors}, where the agent must avoid stepping onto ``frozen'' (unsafe) tiles.
\end{enumerate*}
We discuss the rationale for these two tasks and their importance to imitation learning in Sec.~\ref{sec:env_types}.

\subsection{Structure and Actions} \label{sec:structure}

Labyrinth can create a new structure by specifying the desired number of rows and columns, including height and width. The coordinates of the starting ($s_0$) and goal tiles ($g$) can be either 
\begin{enumerate*}[label=(\roman*)]
    \item user-defined -- the user specifies where the start and goal tiles are located; 
    \item biased -- the starting tile is at the lower-left corner of the labyrinth and the goal is at its upper right, or
    \item unbiased -- the goal and starting tiles are set randomly within the labyrinth, according to a minimum specified distance of each other (cf. the Manhattan distance $d(s_0,g)=\mid x_{s_0} - x_g \mid + \mid y_{s_0} - y_g \mid$).    
\end{enumerate*}
We refer to these as \textit{biased} and \textit{unbiased} due to the nature of the action distribution for all possible solutions in these structures (cf. Sec.~\ref{sec:generalisation}).
Biased structures will maintain the action distribution similar even when switching their structures, while unbiased ones will keep the distributions uniform for all actions, which will require the agent to focus more on each state instead of predicting the most likely actions.

\begin{figure}[h!tb]
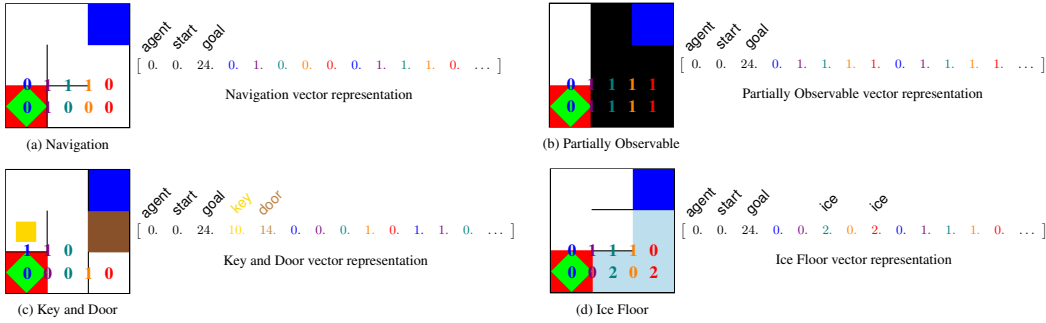

    \centering
    \includestandalone[width=\linewidth]{content/figures/labyrinth3}
    \caption{Different state representations for each task in the Labyrinth.}
    \label{fig:graph}
\end{figure}

It is easier to depict the structure of a labyrinth as a grid, but it is formally defined as a graph, where nodes represent tiles and edges represent connections between them (Figure~\ref{fig:graph}).
The graph we utilise is constructed by removing some edges, which in the visualisation is equivalent to adding a wall sectioning connections between two tiles.
This graph representation allows us to quickly detect duplicates, find all possible solutions between start and goal nodes, and easily create configurations with different degrees of similarity to an existing labyrinth.
Furthermore, configurations can be stored and subsequently reused or altered, allowing for the creation of datasets with specific characteristics and ensuring complete separation between training, validating, and testing sets.

Even though the evaluation and test sets are distinct, we note two different features of this environment:
\begin{enumerate*}[label=(\roman*)]
    \item smaller labyrinths have higher chances of presenting similar trajectories to their goal (different structures while sharing a common path from $s_0$ to $g$; and
    \item the biased setting creates more straightforward solutions since the distribution of actions consists mainly of \up and \rg actions.
\end{enumerate*}

The actions \up, \dn, \rg, and \lf\ move the agent one tile at a time towards the corresponding direction. 
It is important to note that the environment does not prevent an agent from taking an action towards a wall.
However, in such cases, the agent's position will remain the same, although a unit of time will have elapsed.
Finally, no actions can be executed once the goal tile is reached.

\subsection{State and Reward} \label{sec:state_and_reward}

Each state consists of a labyrinth image with the agent drawn in its current position or a vector with the agent, start and goal global positions, and the labyrinth's structure.
The start and goal tiles have different colours, red and blue, respectively, and the agent is a green diamond.
Fig.~\ref{fig:graph}a illustrates the default state of the labyrinth.
Labyrinth can also return a partially observed state. 
This state consists of the tiles and walls in the immediate vicinity of the agent's current position.\footnote{The source code at \url{https://github.com/NathanGavenski/Labyrinth/blob/main/src/labyrinth/utils/render.py\#L157} contains the precise definition of the visibility settings}. 
It is important to note that we do not obfuscate start and goal positions since it would be impossible for the agent to know where these tiles are in the unbiased setting.
In theory, partial observability would make it more complex for the agent to solve the environment.
However, we hypothesise that since imitation learning tries to match the current state with a sample from the teacher, it could make it easier for the agent to reach the goal (even more so when considering the biased setting).
Fig.~\ref{fig:graph}b shows the partial state from the agent's perspective.

Even though imitation learning approaches ignore reward signals from environments in the learning process, we implement a reward function to differentiate the solution for each algorithm in our environment.
For that, we use Eq.~\ref{eq:reward_function}.

\begin{equation} \label{eq:reward_function}
    r_i = 
    \begin{cases}
        \frac{-0.1}{width \times height} & \text{not at goal}  \\ 
        1 + |\tau_s| \times \frac{0.1}{width \times height}  & \text{at goal}
    \end{cases}
\end{equation}

\noindent
In Eq.~\ref{eq:reward_function}, $\mid \tau_s \mid$ is the length of the shortest trajectory.
It allows for the same reward independently of the labyrinth's structure.
In other words, an agent that reaches the goal using the shortest path will always yield an accumulated reward of 1. 
Consequently, an average reward of $1$ means the agent reached the goal in all episodes, which provides a fairer evaluation that is independent of the solution length.
It is important to note that this reward function still gives the agent a positive reward when using a sub-optimal path as long as it does not roam endlessly.
Nevertheless, the use of the reward function is not essential for our experimentation.
If the agent learns to navigate the Labyrinth (i.e, understands how to avoid walls properly and manages to reach the goal in configurations not seen in the training data), we will consider that it has generalised successfully.

\subsection{Settings} \label{sec:env_types}

For the labyrinth, we consider four different settings:
\textit{labyrinth navigation}, where the agent has a typical labyrinth and will need to reach $g$ from $s_0$;
\textit{partially observable labyrinth}, where the agent has to reach the goal but only observes the structure close to it;
\textit{key and door}, where success requires the achievement of sub-goals in a specific sequence.
For example, collect a key from tile $g_k$, before opening a door at tile $g_d$, to then be able to reach the final goal at tile $g$; and
\textit{ice floor}, where the agent must avoid frozen tiles. 

\noindent
\textbf{Labyrinth navigation}: offers a default setting for standard navigation training and evaluation. 
In the user-defined setting, researchers specify the position of $s_0$ and $g$ tiles. 
Alternatively, they can let the environment choose these positions according to the biased or unbiased settings (cf. Sec.~\ref{sec:structure}).
We believe biased settings are more straightforward for imitation learning agents to learn since they will keep the action distributions similar.
Therefore, the agent must only learn to navigate different transition functions from new labyrinths.
On the other hand, a possible evaluation setting for agents is keeping the structure of walls the same and only changing its initial position (same transition function, possibly different action distribution).
Thus, this task allows for training and evaluation:
\begin{enumerate*}[label=(\roman*)]
    \item with different structures but with $s_0$ and $g$ always at the same tiles;
    \item with different structures and with $s_0$ and $g$ in different tiles; and
    \item with the same structures but with $s_0$ and $g$ in different tiles.
\end{enumerate*}

\noindent
\textbf{Partially observable labyrinths}: changes the labyrinth navigation task only to display information close to the agent's position, $s_0$ and $g$ positions.
We believe that using a partially observable environment might help the agent to focus on the relevant information.
By observing the whole structure, the agent might consider a state out-of-distribution when a training sample may be similar from a local perspective.
Using partially observable states does not remove the possibility of evaluating and testing an agent in the same conditions as the navigation setting.
However, we consider changing the structure in an unbiased setting a more complex problem when partial observability is in place since the change in transition function with the out-of-distribution actions leads to a more diverse set of possible solutions.

\noindent
\textbf{Key and door}: setting allows researchers to measure how well their imitation learning agent can learn a sub-task (collecting a key to open a door before reaching a goal).
When creating a labyrinth with the key and door setting, the environment will first define the structure and then find possible positions for the key and door.
To define the environment's structure, we only allow labyrinths with paths that share at least a tile from $s_0$ to $g$.
Doing so avoids instances where the agent could reach $g$ without completing the sub-task.
To define the door's position $g_d$, we find all possible paths from $s_0$ to $g$ and select the last shared tile among them.
To define the key's position $g_k$, we also use all possible paths from $s_0$ to $g$ but select a random reachable tile from all tiles not present in the set.
We select the last shared to ensure the maximum number of tiles possible for the key and select a tile not present in the set of solutions to ensure the agent did not collect the key by chance and that it was an intended decision from the agent.
The key and door setting also allows for the same set of evaluations from the navigation setting with one additional evaluation where we keep the same structure and $s_0$, $g$ and $g_d$ positions and change only $g_k$'s position.

\noindent
\textbf{Ice floor}: offers a setting for researchers to experiment with safety and generalisation problems.
In this setting, if the agent steps on the ice, the tile will break, and the episode will terminate (fail).
For this setting, the environment creates its structure and ensures that at least two possible solutions exist to reach $g$.
We set this premise to guarantee that if we set one possible path with ice floors, there will be at least another path that will be safe for the agent to reach $g$.
With all possible paths, we select one of the possible paths from the set of solutions and set the tiles to be ice.
It is important to note that we only set the tiles unique to that path to avoid accidentally making all paths unsafe. 
During the evaluation, researchers can maintain $s_o$ and $g$ positions and the same structure but swap ice tiles from unsafe to safe paths.

\subsection{Ease of Use, Reproducibility and Customisation} \label{sec:easy_to_use}

We understand that an environment must be easy to use, allow for customisation, and be reproducible for the community to adopt it.
Therefore, we developed Labyrinth with all of these in mind.
Labyrinth runs on `gymnasium'~\cite{}, allowing researchers who already use the highly adopted Python library to use the environment with minimal adaptation.
A typical utilisation of Labyrinth is illustrated below:
\begin{minted}{python}
    import gymnasium as gym
    import labyrinth |\label{code:register}|
    environment = gym.make( |\label{code:env_definition_start}|
        "Labyrinth-v0", shape=(5, 5), occlusion=False,
        key_and_door=False, icy_floor=False, render_mode="rgb_array"
    )|\label{code:env_definition_end}|
    obs, info = environment.reset(options={"agent": True})|\label{code:reset}|
    solutions = environment.solve(mode="all")|\label{code:solver}|
    obs, reward, done, truncated, info = environment.step(action)|\label{code:step}|
\end{minted}
Line~\ref{code:register} registers the environment on gymnasium, Lines~\ref{code:env_definition_start}--\ref{code:env_definition_end} define the environment, Line~\ref{code:reset} creates a new environment and yields the first state, Line~\ref{code:solver} provides all possible solutions for that Labyrinth, and Line~\ref{code:step} performs a random action in the environment, which returns the next state, reward, whether the agent arrived at $g$, whether the agent has fallen through an ice floor, and the environment's info.
To define an instance of the environment the user has the following parameters: $shape$, which requires a tuple that defines the width and height; $occlusion$ sets the partially observable setting; $key\_and\_door$ enables the key and door setting; $icy\_floor$ enables the ice floors setting; and $render\_mode$ defines what type of state should the environment return (vector or image).
It is important to note that $occlusion$, $key\_and\_door$ and $icy\_floor$ are mutually exclusive.
The solver for the environment uses Johnson's algorithm~\cite{johnson1977efficient} to find all possible paths from $s_0$ to $g$.
Beyond all possible solutions, the solver also allows for the shortest solution, which will return a single solution, one of all possible shortest paths (when the structure has more than one path with the same length).

To ensure reproducibility, we allow users to save and load past instances as follows:
\begin{minted}{python}
    from labyrinth.file_utils import convert_from_file, create_file_from_environment
    create_file_from_environment(environment, "example.labyrinth") |\label{code:save}|
    environment.load(*convert_from_file("example.labyrinth")) |\label{code:load}|
\end{minted}

Line~\ref{code:save} saves the current setting of the environment to the file \texttt{example.labyrinth}, and Line~\ref{code:load} loads the file structure and setting in the current Labyrinth object.
Therefore, a user can create a set of structures and settings for training and another for evaluation, keeping consistency between different training and evaluation cycles.
In fact, Labyrinth provides a feature for the easy creation of these sets:
\begin{minted}{bash}
python -m labyrinth.generate --width 5 --height 5 --train 100 --eval 100 --test 100
\end{minted}
where $train$, $eval$ and $test$ define the size of each set ($100$ in this example) and the $width$ and $height$ of the structure.
We reiterate that Labyrinth ensures that each structure is unique by hashing its structure and controlling that each new structure is not present during the creation of all sets, i.e., each structure is unique in its set and among all sets.

\begin{wrapfigure}{l}{0.30\linewidth}
\begin{minted}[escapeinside=@@]{python}
    key_and_lock: False @\label{code:attributes_start}@
    icy_floor: False
    occlusion: False @\label{code:attributes_end}@
    labyrinth: @\label{code:structure_start}@
    -------------  
    |   |     E |
    |   +   + - |
    |           |
    |   + - +   |
    | S |       |
    ------------- 
    end @\label{code:structure_end}@
\end{minted}
\end{wrapfigure}
To allow easy customisation of the environment structure, we create a custom setting language that enables users to visualise the structure of each file easily, but also allows for editing existing structures quickly.
An example of this can be seen here:
where Lines~\ref{code:attributes_start}--\ref{code:attributes_end} define the settings for the environment and Lines~\ref{code:structure_start}--\ref{code:structure_end} defines the structure.
For defining the tile types, users can use $S$ for the first state $s_0$, $E$ for the goal $g$, $K$ and $D$ for key and door positions, respectively, and $I$ for setting ice tile positions.

Finally, we provide a set of labyrinths and data for training imitation learning agents on IL-Datasets~\cite{gavenski2024ildatasets}, which hosts its datasets on HuggingFace~\cite{huggingface2023}.
IL-Datasets provides a convenient and uniform way to evaluate imitation learning methods and ensures that implementations are compared under the same conditions: seeds, training data and evaluation.
Labyrinth can be used without IL-Datasets, it is used for its convenience and the benefits it provides to researchers.
We create datasets for squared labyrinths with sizes of $3$, $4$, and $5$.
Each dataset consists of three splits (train, evaluation, and test), each split consisting of the shortest paths from $s_0$ to $g$ on the biased setting.
Each dataset entry consists of the image observation, action, immediate reward, whether that entry is the first for an episode and the labyrinth information for recreating the same experiment.
If users desire to use the unbiased setting, they can load the information from each entry and change $s_0$ and $g$ positions by using functions $change\_start\_and\_goal$ and $change\_start$.

\section{On the Generalisation Requirements for Benchmarks} \label{sec:generalisation}

For testing generalisation, we believe an environment needs some key requirements:
\begin{enumerate*}[label=(\roman*)]
    \item poses a challenging task;
    \item a significant change from training to evaluation;
    \item controls over these changes; and 
    \item allows for debugging of the agent.
\end{enumerate*}

We argue that the task must be non-trivial for the first requirement and demand reasoning beyond memorisation.
Labyrinth addresses this by requiring agents to plan long-horizon, reason over topological structures, and adapt to altered starting and goal states.
Moreover, its variants, such as key-and-door and icy floor settings, add complexity through temporal dependencies and safety constraints, respectively.
These extensions prevent shortcut solutions and promote learning robust decision-making strategies.
Unlike classical benchmarks, where solutions can often be reduced to reactive policies, solving Labyrinth consistently demands trajectory-level reasoning and adaptation.

For the second requirement, we analyse the Labyrinth environment and the most common environments used in imitation learning benchmarks~\cite{gavenski2024imitation}: MountainCar, CartPole, Hopper, Walker-$2$D and HalfCheetah.
Table~\ref{tab:seeds} shows $100,000$ different initial states for the most common environments.
For it, we initialise the environment with a seed not used for generating the training dataset, and use the closest average distance, based on the Manhattan distance, to it.
We observe that most initial states are quite similar to the training data.
This is not ideal since, by having states that are closer to the training data, imitation learning agents can adopt a behaviour-seeking mode, where the agent tries to use the expert's action instead of predicting the most adequate action for a given state.
Ideally, the reward functions in environments would account for these less-than-optimal actions and show divergence in the behaviour.
However, when doing this analysis, we encounter a significant downside of these environments.
For CartPole and MountainCar, we could reach results comparable to those of the expert by recording a single sequence of expert actions and repeating it in a new initialisation.
For example, for the MountainCar environment, a classical environment with a more challenging dynamic (agents have to build up momentum to reach the goal), we record an episode of accumulated reward of $-106.45$.
By simply using the same sequence of actions over 100 different episodes, we reach an average accumulated reward of $-104.87 \pm 0.8562$.
It is important to note that MountainCar consider the task solved when the agents achieve an average accumulated reward of $-110$.
Yet, classical environments are considered simplistic in nature, as pointed out in Sec.~\ref{sec:intro}.
Therefore, we also analyse how these continuous tasks perform under different initialisations.
In it, we discover that these environments are quite lenient over the actions taking place.
For example, on the Hopper environment, by retrieving the closest state from the current environment one on a different seed and performing the exact expert action from the training data, we achieve a reward of $3530.2367 \pm 15.5748$, while the expert achieves $3536.3626 \pm 9.5699$, a marginally better result.
These results are worrisome since most imitation learning works use these benchmarks to show that their model learned the underlying task and can generalise well to other initialisations.

\begin{table}[t!bh]
\scriptsize
\centering
\caption{Manhattan distance for $1e5$ initialisations for the Gym and DeepMind control suites.}
\label{tab:seeds}
\begin{tabular*}{\textwidth}{l@{\extracolsep{\fill}}rr rr}
\toprule
\multirow{2}{*}{Environment} & \multicolumn{2}{c}{Gym} & \multicolumn{2}{c}{DeepMind} \\
\cmidrule(lr){2-3}
\cmidrule(lr){4-5}

    & \multicolumn{1}{c}{Summation} & \multicolumn{1}{c}{Average} & \multicolumn{1}{c}{Summation} & \multicolumn{1}{c}{Average} \\
\midrule

MountainCar &  $0.0021 \pm 0.0016$ & $0.0010 \pm 0.0070$ &  \multicolumn{1}{c}{-} & \multicolumn{1}{c}{-} \\
CartPole    &  $0.0380 \pm 0.0204$ & $\mathbf{0.0095 \pm 0.0051}$ &  $\mathbf{0.0932 \pm 0.0189}$ & $0.0093 \pm 0.0047$ \\
Hopper      &  $2.3931 \pm 0.0091$ & $0.2175 \pm 0.0008$ &  $\mathbf{3.5610 \pm 0.4974}$ & $\mathbf{0.3237 \pm 0.0452}$ \\
Walker-2D   &  $5.4045 \pm 0.0117$ & $0.3179 \pm 0.0060$ &  $\mathbf{8.3509 \pm 0.6852}$ & $\mathbf{0.4912 \pm 0.0403}$ \\
HalfCheetah & $\mathbf{12.5915 \pm 0.3289}$ & $\mathbf{0.7406 \pm 0.0193}$ & $12.3376 \pm 0.2032$ & $0.7257 \pm 0.0119$ \\ 
\bottomrule

\end{tabular*}
\end{table}

\begin{figure}[t!hb]
    \centering
    \begin{subfigure}{.32\columnwidth}
        \centering
        \includegraphics[width=\linewidth]{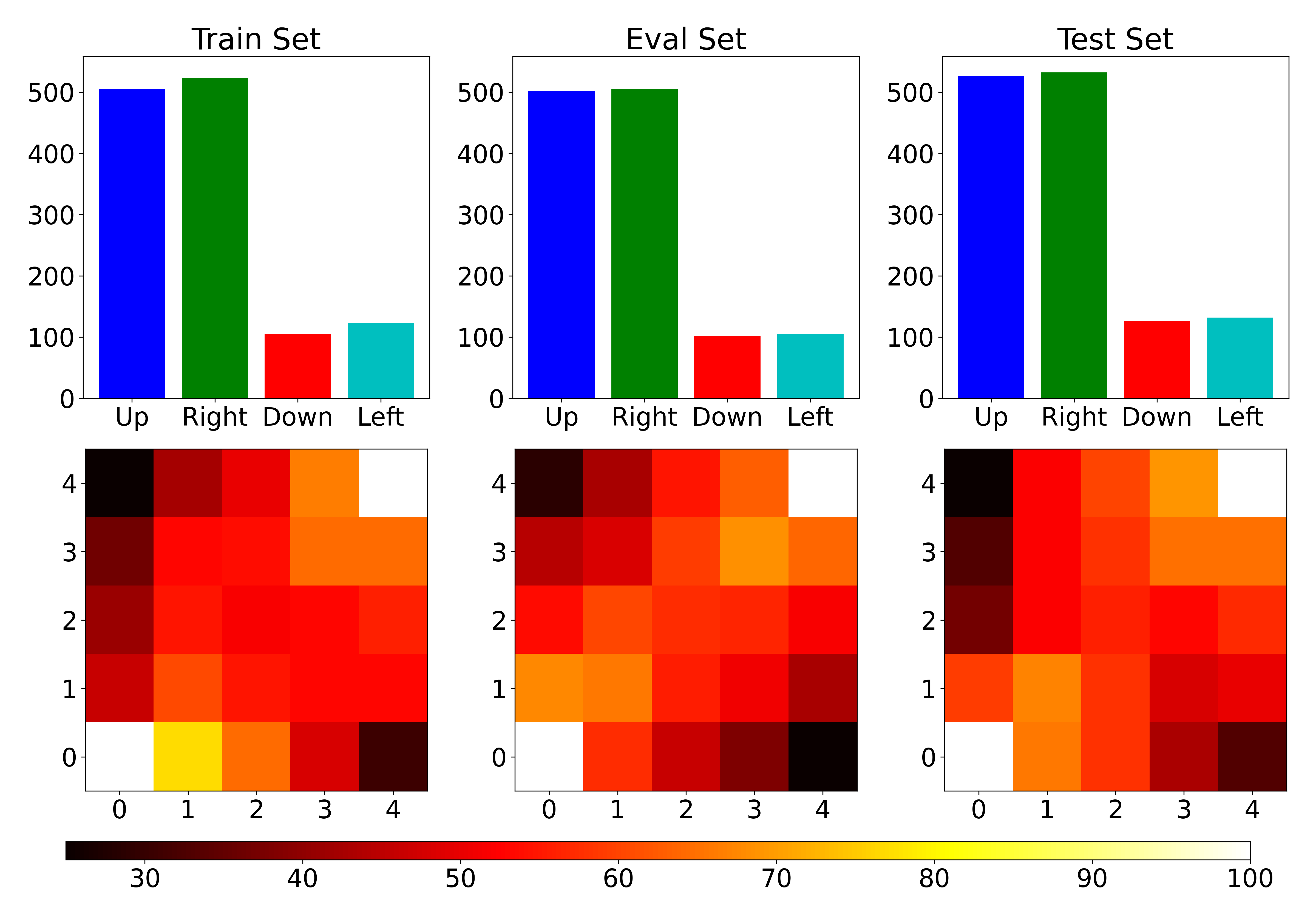}
        \caption{\scriptsize Different structure, same $s_0$ and $g$}
        \label{fig:biased}
    \end{subfigure}
    \hfill
    \begin{subfigure}{.32\textwidth}
        \centering
        \includegraphics[width=\linewidth]{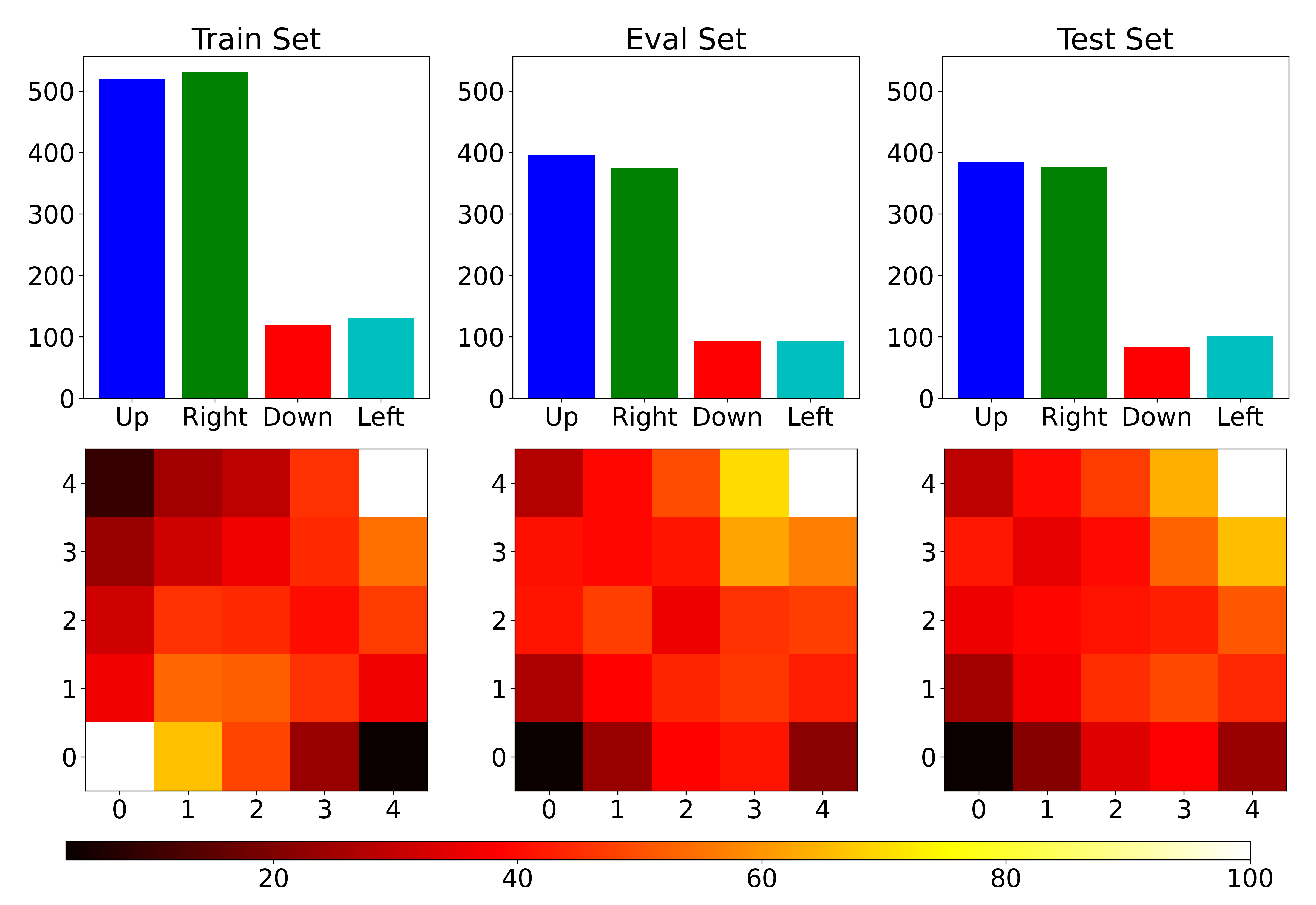}
        \caption{\scriptsize Same structure and $g$, different $s_0$}
    \label{fig:start}
    \end{subfigure}
    \hfill
    \begin{subfigure}{.32\textwidth}
        \centering        
        \includegraphics[width=\linewidth]{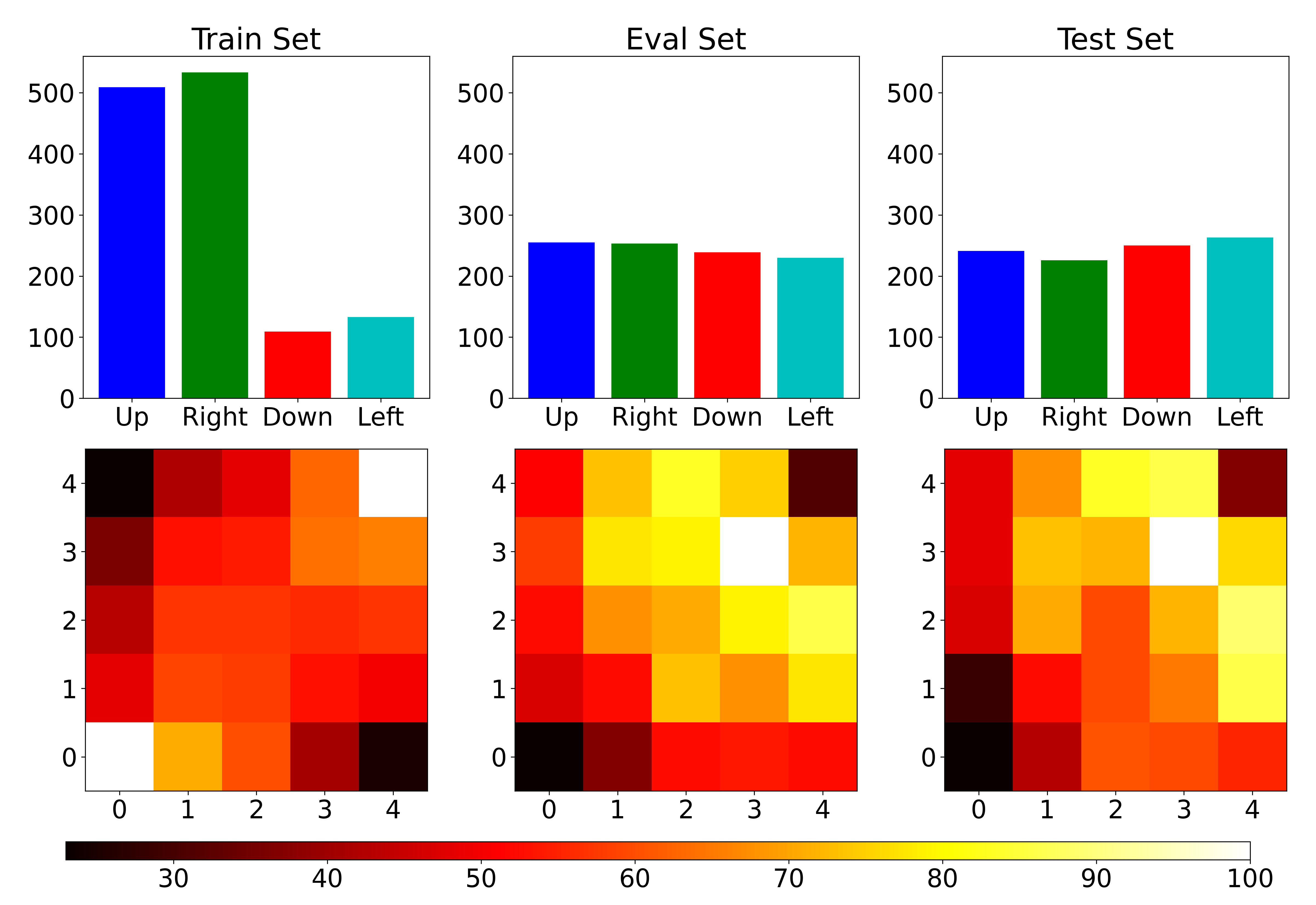}
        \caption{\scriptsize Same structure, different $s_0$ and $g$}
        \label{fig:unbiased}
    \end{subfigure}
    \caption{Tile and action distribution over different settings for the Labyrinth environment.}
    \label{fig:lab-divergence}
\end{figure}

To understand how the labyrinth diverges from training, we analyse the action distribution over all possible settings (described in Sec.~\ref{sec:env_types}).
Figure~\ref{fig:lab-divergence} shows the cell distribution and action distributions for the solutions over the train, evaluation, and test splits for a $5 \times 5$ labyrinth\footnote{The supplementary material contains all other labyrinth sizes.}.
We observe that for the first two settings (Fig.~\ref{fig:biased} and~\ref{fig:start}) the action distribution remains close to the same during each split. 
However, the cell distribution changes, which means that to reach $g$, the agent will have to adapt its solution and better rank information to achieve the goal.
In other words, if the agent only learns to find the closest state to the training data, and perform the same action, there will be labyrinth settings it will not solve.
Moreover, by changing both $s_0$ and $g$ and maintaining the same structure (Fig.~\ref{fig:unbiased}), the action distribution drastically shifts to a more uniform one.

The third requirement is necessary for all methods to be evaluated under fair and explainable conditions.
Labyrinth enables precise control over how environments differ between training and evaluation, allowing researchers to isolate specific generalisation challenges.
For instance, one can hold the labyrinth structure fixed while varying the agent's initial position (Fig.~\ref{fig:start}), or conversely, alter the structure while maintaining consistent start and goal locations (Fig.~\ref{fig:biased}).
This granularity helps identify whether failure modes are due to perceptual mismatch, action distribution shifts, or poor task abstraction.
Full access to the graph structure makes it easier to inspect agent failures and identify brittle behaviour, which is often opaque in high-dimensional or continuous control settings.

The final requirement is that the environment must allow for effective debugging and inspection of the agent's behaviour.
Labyrinth satisfies this by offering full access to both the structural and observational components of the environment, and by allowing researchers to place the agent in any arbitrary state.
More importantly, due to its discrete and fully defined transition graph, Labyrinth allows us to compute the optimal action for every individual state under any configuration.
This enables researchers to directly test whether the agent selects the correct action in a given state, quantify deviations from optimal behaviour, and compare across structurally similar settings.
In contrast, widely used benchmarks such as MuJoCo-based (e.g., Hopper, HalfCheetah) and Atari-based environments make it virtually impossible to define the optimal action in most states due to high-dimensional, continuous dynamics and implicit goals. Similarly, in visual environments like Atari, researchers may not even have access to the full internal state, making it difficult to determine what constitutes a correct action.
Labyrinth's explicit structure and ground-truth optimality afford a level of transparency and controllability that these environments lack, making it especially suitable for interpretability, policy debugging, and fine-grained evaluation of generalisation.

\section{Benchmarking common imitation learning methods}\label{sec:experiments}

We now benchmark some imitation learning methods to demonstrate the effectiveness of this environment in testing generalisation.
Due to space constraints, this section only displays the results for the \textit{labyrinth navigation} setting. 
We show all other settings in the supplementary material.

\subsection{Implementations and Metrics}

We use implementations from IL-Datasets~\cite{gavenski2024ildatasets} for all imitation learning methods.
IL-Datasets provides us with implementations for 
Behavioural Cloning (BC)~\cite{Pomerleau1988}, 
DAgger~\cite{ross2011reduction}, 
Generative Adversarial Imitation Learning (GAIL)~\cite{ho2016generative}, 
Behavioural Cloning from Observation (BCO)~\cite{TorabiEtAl2018}, 
Soft Q Imitation Learning (SQIL)~\cite{reddy2019sqil}, and
Imitating Unknown Policies via Exploration (IUPE)~\cite{gavenski2020iupe}. 
We selected these methods because they offer a diverse range of imitation learning approaches.
BC, GAIL, DAgger, and SQIL are all imitation learning from demonstration methods, while BCO and IUPE are imitation learning from observation methods.
Moreover, DAgger requires access to the expert, which can benefit the training since any given labyrinth knows the optimal action.
On the other hand, GAIL, BC and BCO are offline (do not interact with the environment during training), and the others are online.
To prevent any method from accessing other labyrinth structures outside of those in the training data, we enforce that the online portion of their training is conducted only under the same conditions as those in which the dataset was created.
In other words, we load the same labyrinth structures from the training dataset split during these interactions.
Finally, they are also diverse in their learning approaches, employing adversarial and inverse reinforcement learning (GAIL, DAgger, and SQIL), dynamic methods~\cite{gavenski2024imitation} (BCO and IUPE), or behavioural cloning (BC).

In this work, we use two metrics: \textit{average episodic reward} ($AER$) and \textit{success ratio} ($SR$).
$AER$ is the average reward the agent accumulates over $n$ episodes.
In our experiments, we display the $AER$ for each of the $100$ train, evaluation, and test labyrinths.
An $AER$ of $1$ means the agent achieves the goal using the shortest path.
$SR$ is the ratio of the agent achieving the goal tile over $n$ episodes.

\subsection{Results}

Table~\ref{tab:benchmark} shows the benchmark results for each method in a $5 \times 5$ labyrinth with the same starting and goal tiles (biased setting) for the training, evaluation, and test splits.
The dataset for this benchmark\footnote{\url{https://huggingface.co/datasets/NathanGavenski/Labyrinth-v0_5x5}} (and for all others in the supplementary material) is hosted on HuggingFace~\cite{huggingface2023}, as explained in Section~\ref{sec:easy_to_use}, and we provide all links to the datasets used in this work in the supplementary material.
Besides the images, the dataset also contains all the information needed to recreate each labyrinth according to each entry.
The experiments in Tab.~\ref{tab:benchmark} use the convolutional neural network based on the original Atari Deep Q-Network~\cite{mnih2013playing} as the encoder for each model, and we train all methods for $1,000$ epochs. 
As an addendum, we conducted additional experiments for other labyrinth sizes and settings, and a brief ablation of other neural network structures.
These are described in the supplemental material.

Our experiments show that pure imitation learning methods (those not using inverse reinforcement learning techniques) perform better in Labyrinth. 
We believe that the reason for this is that pure imitation learning methods rely primarily on supervised learning losses, which encourages these models to learn better encodings for each image state.
This results in the model generalising more, i.e., performing better in labyrinth structures not seen during training.
When looking for the closest training examples in the encoding space given an evaluation or a test input, we discover that the agent's position itself for these models is less important than the actual wall structure surrounding the position.
In these cases, the closest training images might have the agent in a different position, but the wall structure remains similar.
The inverse reinforcement learning methods' optimisation is less direct, and the models do not learn the same patterns, resulting in less optimal behaviour.
Unfortunately, all methods perform poorly in this setting, except IUPE, which is the only method to achieve a result higher than $10\%$ on the evaluation set.
Yet, this result did not translate into the test split, which we see as evidence that IUPE did not learn the navigation task itself. It generalised well in the validation set, but not in the testing one. The other methods performed similarly badly in the test and validation sets.

\begin{table}[th]
    \scriptsize
    \centering
    \caption{Benchmark results for training, validation and testing splits.}
    \label{tab:benchmark}
    \begin{tabular*}{\textwidth}{@{\extracolsep{\fill}} lc r r r r r r}
        \toprule
        Splits & Metric
            & \multicolumn{1}{c}{BC}  
            & \multicolumn{1}{c}{DAgger}
            & \multicolumn{1}{c}{GAIL}  
            & \multicolumn{1}{c}{BCO}  
            & \multicolumn{1}{c}{SQIL}  
            & \multicolumn{1}{c}{IUPE} \\ \midrule
        \multirow{2}{*}{Train} & $AER$ & $-2.11 \pm 2.41$ & $-1.18 \pm 2.45$ & $-0.98 \pm 1.89$ & $-0.53 \pm 2.23$ & $-3.80 \pm 0.96$ & $\mathbf{0.27 \pm 2.39}$ \\
            & $SR$ & $37\%$ & $57\%$ & $61\%$ & $70\%$ & $4\%$ & $\mathbf{75\%}$ \\
        \multirow{2}{*}{Valid.} & $AER$ & $-3.70 \pm 1.18$ & $-3.75 \pm 1.08$ & $-3.57 \pm 1.58$ & $-3.90 \pm 0.69$ & $-3.95 \pm 0.49$ & $\mathbf{-2.80 \pm 2.12}$ \\
            & $SR$ & $6\%$ & $5\%$ & $9\%$ & $2\%$ & $1\%$ & $\mathbf{21\%}$ \\
        \multirow{2}{*}{Test} & $AER$ & $-3.90 \pm 0.70$ & $-3.80 \pm 0.97$ & $-3.85 \pm 0.85$ & $-3.85 \pm 0.85$ & $-4.00 \pm 0.00$ & $\mathbf{-3.85 \pm 1.00}$ \\
            & $SR$ & $2\%$ &$4\%$ & $3\%$ & $3\%$ & $0\%$ & $\mathbf{5\%}$ \\
        \bottomrule
    \end{tabular*}
\end{table}

Finally, to understand whether Labyrinth was too complex a challenge for imitation learning, we evaluated BC under an extended period of training ($10,000$ epochs) and using a more robust neural network architecture (ResNet-18~\cite{HeEtAl2016}).
We chose BC as the baseline for this experiment because it is the most simplistic approach to pure imitation learning and the worst-performing of these methods.
When running BC with the same structure but for an extended period, it achieved $100\%$ during training, $41\%$ in the validation, and $34\%$ in the test splits, an improvement over Tab.~\ref{tab:benchmark} results.
Yet, when running BC with a ResNet encoder, it achieves $100\%$ $SR$ in the training, $56\%$ in the validation and $53\%$ in the test splits, a significant improvement from the results in Tab.~\ref{tab:benchmark}.
We believe these numbers result from the model learning a less spurious encoding space.
Therefore, with time, the model learns the correct characteristics to classify the correct action.
However, it does not learn how to perform the underlying task of navigating the structure.
This is backed up by the fact that improving the encoding architecture improves the method's performance, but does not guarantee the results from the other splits.

In summary, our experiments highlight that existing imitation learning methods struggle to generalise effectively in the Labyrinth environment, especially when faced with unseen structures. 
These results show Labyrinth's suitability for rigorous generalisation testing and underscore the need for more robust learning approaches.

\section{Conclusion}\label{sec:conclusion}

In this work, we proposed Labyrinth, an easy-to-use, reproducible, and customisable environment for testing generalisation with imitation learning agents.
Labyrinth provides researchers with:
\begin{enumerate*}[label=(\roman*)]
    \item a way to explicitly separate training, validation and test data via different labyrinth structures, start and goal positions;
    \item a discrete and fully observable state space where all possible states, transitions and optimal actions are explicitly defined, enabling precise analysis of an agent's decision-making process;
    \item a way to systematically analyse an agent's ability to learn and adapt to structural changes and action distribution shifts, offering insights into the agent's robustness and generalisation capabilities; and
    \item the ability to increase difficulty while preserving the nature of the task, and to analyse the inner parameters of the agents.
\end{enumerate*}

We analysed other commonly used imitation learning benchmarks and showed how the field could benefit from using Labyrinth as a platform for testing generalisation.
Labyrinth is challenging enough to require agents to learn the underlying task to solve each unseen labyrinth structure.
It offers customisable evaluation sets that are different enough from the training data (e.g., action distribution shift and other transition functions) to allow for controlled evaluation and debugging of each agent.
Furthermore, Labyrinth provides the same features as all other standard benchmarks, such as accessibility via gymnasium, vector and image representations, and a reward function to compare different agents' results.

We performed a benchmark in the Labyrinth environment using common imitation learning methods, concluding that the field has yet to improve its generalisation capabilities.
Although machine learning techniques can improve their results when solving unseen structures, they still do not generalise well, even if the action distribution remains the same.
Moreover, the type of generalisation from the machine learning field would not theoretically apply to the required generalisation for the agents in this setting.
To achieve a high success rate across each split, agents must build knowledge for the underlying task (navigation) instead of only correlating training samples to the agent's current state.
We believe Labyrinth can help researchers benchmark their method's generalisation capabilities and improve the field perception over how to benchmark novel methods better.

Finally, Labyrinth comes with some limitations we envision tackling in the future.
As it is developed now, Labyrinth only allows for discrete actions, which is ideal for finding the optimal action for each state.
However, some imitation learning methods are only suited for continuous actions, such as OPOLO~\cite{zhu2020opolo}, MAHALO~\cite{li2023mahalo} and CILO~\cite{gavenski2024explorative}.
Ideally, we would like to provide the option of performing continuous actions while keeping all the features Labyrinth provides (cf.~\ref{sec:environment}), which other labyrinth-like environments do not have (such as Ant and Point Maze~\cite{gymnasium_robotics2023github}).
We would also like to develop a customisable tile feature that would allow researchers to specify particular behaviours in some tiles easily.
As it stands now, this can be done, but requires researchers to change the source code in the environment. 

\newpage
\begin{ack}
This work was supported by UK Research and Innovation [grant number EP/S023356/1], in the UKRI Centre for Doctoral Training in Safe and Trusted Artificial Intelligence (\url{www.safeandtrustedai.org}) and made possible via King's Computational Research, Engineering and Technology Environment (CREATE)~\cite{create}.
\end{ack}

\bibliographystyle{plainnat}
\bibliography{reference}

\begin{thebibliography}{28}
\providecommand{\natexlab}[1]{#1}
\providecommand{\url}[1]{\texttt{#1}}
\expandafter\ifx\csname urlstyle\endcsname\relax
  \providecommand{\doi}[1]{doi: #1}\else
  \providecommand{\doi}{doi: \begingroup \urlstyle{rm}\Url}\fi

\bibitem[Aytar et~al.(2018)Aytar, Pfaff, Budden, Paine, Wang, and
  de~Freitas]{aytar2018playing}
Yusuf Aytar, Tobias Pfaff, David Budden, Thomas Paine, Ziyu Wang, and Nando
  de~Freitas.
\newblock Playing hard exploration games by watching youtube.
\newblock In \emph{Advances in Neural Information Processing Systems}, pages
  2930--2941. Advances in Neural Information Processing Systems, 2018.

\bibitem[Barto et~al.(1983)Barto, Sutton, and Anderson]{barto1983neuronlike}
Andrew~G Barto, Richard~S Sutton, and Charles~W Anderson.
\newblock Neuronlike adaptive elements that can solve difficult learning
  control problems.
\newblock \emph{IEEE transactions on systems, man, and cybernetics}, 1\penalty0
  (5):\penalty0 834--846, 1983.

\bibitem[Chevalier-Boisvert et~al.(2023)Chevalier-Boisvert, Dai, Towers,
  Perez-Vicente, Willems, Lahlou, Pal, Castro, and
  Terry]{chevalier2023minigrid}
Maxime Chevalier-Boisvert, Bolun Dai, Mark Towers, Rodrigo Perez-Vicente, Lucas
  Willems, Salem Lahlou, Suman Pal, Pablo~Samuel Castro, and Jordan Terry.
\newblock Minigrid \& miniworld: Modular \& customizable reinforcement learning
  environments for goal-oriented tasks.
\newblock \emph{Advances in Neural Information Processing Systems},
  36:\penalty0 73383--73394, 2023.

\bibitem[Cobbe et~al.(2020)Cobbe, Hesse, Hilton, and
  Schulman]{cobbe2020leveraging}
Karl Cobbe, Chris Hesse, Jacob Hilton, and John Schulman.
\newblock Leveraging procedural generation to benchmark reinforcement learning.
\newblock In \emph{International conference on machine learning}, pages
  2048--2056. PMLR, 2020.

\bibitem[de~Lazcano et~al.(2024)de~Lazcano, Andreas, Tai, Lee, and
  Terry]{gymnasium_robotics2023github}
Rodrigo de~Lazcano, Kallinteris Andreas, Jun~Jet Tai, Seungjae~Ryan Lee, and
  Jordan Terry.
\newblock Gymnasium robotics, 2024.
\newblock URL \url{http://github.com/Farama-Foundation/Gymnasium-Robotics}.

\bibitem[e~Research~team(2023)]{create}
King's College~London e~Research~team.
\newblock King's computational research, engineering and technology environment
  (create), 2023.
\newblock URL \url{https://doi.org/10.18742/rnvf-m076}.

\bibitem[Face(2023)]{huggingface2023}
Hugging Face.
\newblock Hugging face.
\newblock Web Page, 2023.
\newblock URL \url{https://huggingface.co/}.

\bibitem[Gavenski et~al.(2020)Gavenski, Monteiro, Granada, Meneguzzi, and
  Barros]{gavenski2020iupe}
Nathan Gavenski, Juarez Monteiro, Roger Granada, Felipe Meneguzzi, and
  Rodrigo~C Barros.
\newblock Imitating unknown policies via exploration.
\newblock In \emph{Proceedings of the 2020 British Machine Vision Virtual
  Conference}, BMVC 2020, pages 1--8. Proceedings of the 2020 British Machine
  Vision Virtual Conference, BMVA, 2020.

\bibitem[Gavenski et~al.(2024{\natexlab{a}})Gavenski, Luck, and
  Rodrigues]{gavenski2024ildatasets}
Nathan Gavenski, Michael Luck, and Odinaldo Rodrigues.
\newblock Imitation learning datasets: A toolkit for creating datasets,
  training agents and benchmarking.
\newblock In \emph{Proceedings of the 23rd International Conference on
  Autonomous Agents and Multiagent Systems}, AAMAS '24, page 2800–2802,
  Richland, SC, 2024{\natexlab{a}}. International Foundation for Autonomous
  Agents and Multiagent Systems.
\newblock ISBN 9798400704864.

\bibitem[Gavenski et~al.(2024{\natexlab{b}})Gavenski, Monteiro, Meneguzzi,
  Luck, and Rodrigues]{gavenski2024explorative}
Nathan Gavenski, Juarez Monteiro, Felipe Meneguzzi, Michael Luck, and Odinaldo
  Rodrigues.
\newblock Explorative imitation learning: A path signature approach for
  continuous environments.
\newblock In \emph{ECAI 2024}, pages 1551--1558. IOS Press, 2024{\natexlab{b}}.

\bibitem[Gavenski et~al.(2024{\natexlab{c}})Gavenski, Rodrigues, and
  Luck]{gavenski2024imitation}
Nathan Gavenski, Odinaldo Rodrigues, and Michael Luck.
\newblock Imitation learning: a survey of learning methods, environments and
  metrics.
\newblock \emph{arXiv e-prints}, pages arXiv--2404, 2024{\natexlab{c}}.

\bibitem[Hafner(2021)]{hafner2021benchmarking}
Danijar Hafner.
\newblock Benchmarking the spectrum of agent capabilities.
\newblock \emph{arXiv preprint arXiv:2109.06780}, 2021.

\bibitem[He et~al.(2016)He, Zhang, Ren, and Sun]{HeEtAl2016}
Kaiming He, Xiangyu Zhang, Shaoqing Ren, and Jian Sun.
\newblock Deep residual learning for image recognition.
\newblock In \emph{Proceedings of the 2016 IEEE Conference on Computer Vision
  and Pattern Recognition}, CVPR 2016, pages 770--778. Proceedings of the 2016
  IEEE Conference on Computer Vision and Pattern Recognition, 2016.

\bibitem[Ho and Ermon(2016)]{ho2016generative}
Jonathan Ho and Stefano Ermon.
\newblock Generative adversarial imitation learning.
\newblock In \emph{Advances in neural information processing systems}, pages
  4565--4573. Advances in neural information processing systems, 2016.

\bibitem[Johnson(1977)]{johnson1977efficient}
Donald~B Johnson.
\newblock Efficient algorithms for shortest paths in sparse networks.
\newblock \emph{Journal of the ACM (JACM)}, 24\penalty0 (1):\penalty0 1--13,
  1977.

\bibitem[Li et~al.(2023)Li, Boots, and Cheng]{li2023mahalo}
Anqi Li, Byron Boots, and Ching-An Cheng.
\newblock Mahalo: Unifying offline reinforcement learning and imitation
  learning from observations.
\newblock In \emph{International Conference on Machine Learning}. PMLR, 2023.

\bibitem[McLean et~al.(2025)McLean, Chatzaroulas, McCutcheon, R{\"o}der, Yu,
  He, Zentner, Julian, Terry, Woungang, et~al.]{metaworld}
Reginald McLean, Evangelos Chatzaroulas, Luc McCutcheon, Frank R{\"o}der,
  Tianhe Yu, Zhanpeng He, KR~Zentner, Ryan Julian, JK~Terry, Isaac Woungang,
  et~al.
\newblock Meta-world+: An improved, standardized, rl benchmark.
\newblock \emph{arXiv preprint arXiv:2505.11289}, 2025.

\bibitem[Mnih et~al.(2013)Mnih, Kavukcuoglu, Silver, Graves, Antonoglou,
  Wierstra, and Riedmiller]{mnih2013playing}
Volodymyr Mnih, Koray Kavukcuoglu, David Silver, Alex Graves, Ioannis
  Antonoglou, Daan Wierstra, and Martin Riedmiller.
\newblock Playing atari with deep reinforcement learning.
\newblock \emph{arXiv preprint arXiv:1312.5602}, 2013.

\bibitem[Moore(1990)]{moore1990efficient}
Andrew~William Moore.
\newblock \emph{Efficient memory-based learning for robot control}.
\newblock PhD thesis, University of Cambridge, 1990.

\bibitem[Pomerleau(1988)]{Pomerleau1988}
Dean~A. Pomerleau.
\newblock Alvinn: An autonomous land vehicle in a neural network.
\newblock In \emph{Proceedings of the 1st Conference on Neural Information
  Processing Systems}, NIPS 1988, pages 305--313. Proceedings of the 1st
  Conference on Neural Information Processing Systems, 1988.

\bibitem[Reddy et~al.(2019)Reddy, Dragan, and Levine]{reddy2019sqil}
Siddharth Reddy, Anca~D Dragan, and Sergey Levine.
\newblock Sqil: Imitation learning via reinforcement learning with sparse
  rewards.
\newblock \emph{arXiv preprint arXiv:1905.11108}, 2019.

\bibitem[Ross et~al.(2011)Ross, Gordon, and Bagnell]{ross2011reduction}
St{\'e}phane Ross, Geoffrey Gordon, and Drew Bagnell.
\newblock A reduction of imitation learning and structured prediction to
  no-regret online learning.
\newblock In \emph{Proceedings of the fourteenth international conference on
  artificial intelligence and statistics}, pages 627--635. JMLR Workshop and
  Conference Proceedings, 2011.

\bibitem[Schulman et~al.(2015)Schulman, Moritz, Levine, Jordan, and
  Abbeel]{schulman2015high}
John Schulman, Philipp Moritz, Sergey Levine, Michael Jordan, and Pieter
  Abbeel.
\newblock High-dimensional continuous control using generalized advantage
  estimation.
\newblock \emph{arXiv preprint arXiv:1506.02438}, 2015.

\bibitem[Tassa et~al.(2018)Tassa, Doron, Muldal, Erez, Li, Casas, Budden,
  Abdolmaleki, Merel, Lefrancq, et~al.]{tassa2018deepmind}
Yuval Tassa, Yotam Doron, Alistair Muldal, Tom Erez, Yazhe Li, Diego de~Las
  Casas, David Budden, Abbas Abdolmaleki, Josh Merel, Andrew Lefrancq, et~al.
\newblock Deepmind control suite.
\newblock \emph{arXiv preprint arXiv:1801.00690}, 1:\penalty0 1--24, 2018.

\bibitem[Torabi et~al.(2018)Torabi, Warnell, and Stone]{TorabiEtAl2018}
Faraz Torabi, Garrett Warnell, and Peter Stone.
\newblock Behavioral cloning from observation.
\newblock In \emph{Proceedings of the 27th International Joint Conference on
  Artificial Intelligence}, pages 4950--4957. Proceedings of the 27th
  International Joint Conference on Artificial Intelligence, 2018.

\bibitem[Wan et~al.(2024)Wan, Wang, Wang, Erickson, and Held]{difftori}
Weikang Wan, Ziyu Wang, Yufei Wang, Zackory Erickson, and David Held.
\newblock Difftori: Differentiable trajectory optimization for deep
  reinforcement and imitation learning.
\newblock \emph{Advances in Neural Information Processing Systems},
  37:\penalty0 109430--109459, 2024.

\bibitem[Wei et~al.(2023)Wei, Sun, Zheng, Vemprala, Bonatti, Chen, Madaan, Ba,
  Kapoor, and Ma]{isimitation}
Yao Wei, Yanchao Sun, Ruijie Zheng, Sai Vemprala, Rogerio Bonatti, Shuhang
  Chen, Ratnesh Madaan, Zhongjie Ba, Ashish Kapoor, and Shuang Ma.
\newblock Is imitation all you need? generalized decision-making with
  dual-phase training.
\newblock In \emph{Proceedings of the IEEE/CVF International Conference on
  Computer Vision}, pages 16221--16231, 2023.

\bibitem[Zhu et~al.(2020)Zhu, Lin, Dai, and Zhou]{zhu2020opolo}
Zhuangdi Zhu, Kaixiang Lin, Bo~Dai, and Jiayu Zhou.
\newblock Off-policy imitation learning from observations.
\newblock In H.~Larochelle, M.~Ranzato, R.~Hadsell, M.F. Balcan, and H.~Lin,
  editors, \emph{Advances in Neural Information Processing Systems}, volume~33,
  pages 12402--12413. Curran Associates, Inc., 2020.
\newblock URL
  \url{https://proceedings.neurips.cc/paper/2020/file/92977ae4d2ba21425a59afb269c2a14e-Paper.pdf}.

\end{thebibliography}

\newpage
\section*{NeurIPS Paper Checklist}

\begin{enumerate}

\item {\bf Claims}
    \item[] Question: Do the main claims made in the abstract and introduction accurately reflect the paper's contributions and scope?
    \item[] Answer: \answerYes{} 
    \item[] Justification: We provide all evidence for the claims made in the abstract.

\item {\bf Limitations}
    \item[] Question: Does the paper discuss the limitations of the work performed by the authors?
    \item[] Answer: \answerYes{} 
    \item[] Justification: In Section~\ref{sec:conclusion}, we discuss all limitations of the proposed environment and future work.

\item {\bf Theory assumptions and proofs}
    \item[] Question: For each theoretical result, does the paper provide the full set of assumptions and a complete (and correct) proof?
    \item[] Answer: \answerNA{} 
    \item[] Justification: This work does not introduce any new theories or require any proofs.

    \item {\bf Experimental result reproducibility}
    \item[] Question: Does the paper fully disclose all the information needed to reproduce the main experimental results of the paper to the extent that it affects the main claims and/or conclusions of the paper (regardless of whether the code and data are provided or not)?
    \item[] Answer: \answerYes{} 
    \item[] Justification: We provide the code for the environment, which can be installed using the GitHub link, the datasets are hosted on HuggingFace with the croissant files provided (as requested for all submissions), and the baselines are all provided by the IL-Datasets package, which allows for running the benchmarks with the same parameters.

\item {\bf Open access to data and code}
    \item[] Question: Does the paper provide open access to the data and code, with sufficient instructions to faithfully reproduce the main experimental results, as described in supplemental material?
    \item[] Answer: \answerYes{} 
    \item[] Justification: We provide all links to the datasets, the code for the environment and the baselines.

\item {\bf Experimental setting/details}
    \item[] Question: Does the paper specify all the training and test details (e.g., data splits, hyperparameters, how they were chosen, type of optimizer, etc.) necessary to understand the results?
    \item[] Answer: \answerYes{} 
    \item[] Justification: The supplementary material contains all information for reproducibility, with the learning rates used on IL-Datasets and the splits provided from the HuggingFace dataset.

\item {\bf Experiment statistical significance}
    \item[] Question: Does the paper report error bars suitably and correctly defined or other appropriate information about the statistical significance of the experiments?
    \item[] Answer: \answerYes{} 
    \item[] Justification: Both tables~\ref{tab:seeds} and~\ref{tab:benchmark} provide their error margins via standard deviation.

\item {\bf Experiments compute resources}
    \item[] Question: For each experiment, does the paper provide sufficient information on the computer resources (type of compute workers, memory, time of execution) needed to reproduce the experiments?
    \item[] Answer: \answerYes{} 
    \item[] Justification: The supplementary material describes the hardware used for experimentation.
    
\item {\bf Code of ethics}
    \item[] Question: Does the research conducted in the paper conform, in every respect, with the NeurIPS Code of Ethics \url{https://neurips.cc/public/EthicsGuidelines}?
    \item[] Answer: \answerYes{} 
    \item[] Justification: We follow the code of ethics from NeurIPS.

\item {\bf Broader impacts}
    \item[] Question: Does the paper discuss both potential positive societal impacts and negative societal impacts of the work performed?
    \item[] Answer: \answerNA{} 
    \item[] Justification: Although we belive this work will have a positive impact to the imitation learning community, we do not expect any societal impacts.
    
\item {\bf Safeguards}
    \item[] Question: Does the paper describe safeguards that have been put in place for responsible release of data or models that have a high risk for misuse (e.g., pretrained language models, image generators, or scraped datasets)?
    \item[] Answer: \answerNA{} 
    \item[] Justification: The environment does not uses any data or models that have high risk for misuse and is all based on public data (coded implementations for all baselines).

\item {\bf Licenses for existing assets}
    \item[] Question: Are the creators or original owners of assets (e.g., code, data, models), used in the paper, properly credited and are the license and terms of use explicitly mentioned and properly respected?
    \item[] Answer: \answerNA{} 
    \item[] Justification: All assets used in this work were created by the authors.

\item {\bf New assets}
    \item[] Question: Are new assets introduced in the paper well documented and is the documentation provided alongside the assets?
    \item[] Answer: \answerYes{} 
    \item[] Justification: We follow the best practices for python implementations, using unit tests and code documentation to guarantee the quality of the environment. We follow the IL-Datasets pattern for the dataset and document the structure on the HuggingFace page.

\item {\bf Crowdsourcing and research with human subjects}
    \item[] Question: For crowdsourcing experiments and research with human subjects, does the paper include the full text of instructions given to participants and screenshots, if applicable, as well as details about compensation (if any)? 
    \item[] Answer: \answerNA{} 
    \item[] Justification: This research does not involve any crowdsourcing or research with human subjects.

\item {\bf Institutional review board (IRB) approvals or equivalent for research with human subjects}
    \item[] Question: Does the paper describe potential risks incurred by study participants, whether such risks were disclosed to the subjects, and whether Institutional Review Board (IRB) approvals (or an equivalent approval/review based on the requirements of your country or institution) were obtained?
    \item[] Answer: \answerNA{} 
    \item[] Justification: This research does not involve any crowdsourcing or research with human subjects.

\item {\bf Declaration of LLM usage}
    \item[] Question: Does the paper describe the usage of LLMs if it is an important, original, or non-standard component of the core methods in this research? Note that if the LLM is used only for writing, editing, or formatting purposes and does not impact the core methodology, scientific rigorousness, or originality of the research, declaration is not required.
    \item[] Answer: \answerNA{} 
    \item[] Justification: We do not use LLMs for any part of this work. Neither the implementation, the dataset creation, nor the writing used LLMs.

\end{enumerate}

\appendix
\section{Datasets}

Labyrinth provides three datasets, for three different sizes: (i) $3 \times 3$; (ii) $4 \times 4$; and (iii) $5 \times 5$.
All datasets consist of at least one solution for $100$ different labyrinths, with the exception of $3 \times 3$, which has $32$ labyrinths per split.
Each split is entirely unique, and no labyrinth appears more than once in the entire dataset, regardless of its split.
Each entry consists of five pieces of information:
\begin{itemize}
    \item \textit{obs}: the path to the image representation for that state;
    \item \textit{actions}: the integer action performed for that solution in that state;
    \item \textit{rewards}:  the float reward received for that action in that state;
    \item \textit{episode\_starts}: the boolean status that states if it is the first state in an episode; and
    \item \textit{info}: the textual information required to load the same labyrinth structure if needed.
\end{itemize}
We provide images in each dataset since we believe that visual information is more useful to the imitation learning agent, and if a vector representation is needed, the \textit{info} parameter allows researchers to load the same structure and the \textit{actions} enables the recreation of the dataset in its vector format.

Each observation is an image of size $600 \times 600 \times 3$.
Although each baseline trained in this work uses a $64 \times 64 \times 3$ input, we thought that providing a bigger image would benefit models requiring downsizing (e.g., the walls will not disappear during resizing).
We note that the dataset has the last state in each episode (when the agent is at the goal $g$). 
Therefore, some entries might not be helpful to the agent.
We recommend researchers to split entries intro episodes (since they are in sequence), and create tuples ($s$, $a$) or ($s$, $a$, $s'$), where $s$ is the state, $a$ is the action, and $s'$ is the state resulting from the transition function given $s$ and $a$.
IL-Datasets'~\cite{gavenski2024ildatasets} `BaselineDataset' provides this implementation as an example.
All datasets can be found in the HuggingFace collection at: \url{https://huggingface.co/collections/NathanGavenski/labyrinth-datasets-68245a55019fed6983502805}.

\section{Configuration Language}

As explained in Sec.~\ref{sec:easy_to_use} of the main work, Labyrinth uses its own configuration language to save and load existing structures.
We create this language to support researchers when manually editing a labyrinth.
Initially, we stored each labyrinth by writing the edges of the graph that models the structure (cf. Sec.~\ref{sec:structure} of the main work).
However, modifying the file required a lot of effort since edges are stored by their global location (the inline position instead of their $x$ and $y$ coordinates).
Therefore, we envisioned a more friendly visualisation that would help researchers modify the structure or the task without the need for a deep understanding of the labyrinth's inner structure.

\begin{figure}[h!tb]
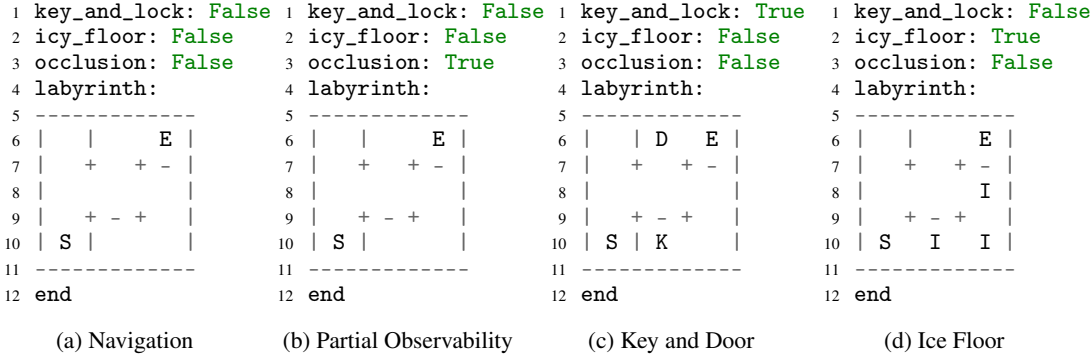

    \begin{subfigure}[b]{0.22\textwidth}
       \centering
\begin{minted}[numbersep=-4pt, escapeinside=@@]{python}
  key_and_lock: False
  icy_floor: False
  occlusion: False
  labyrinth:
  -------------  
  |   |     E |
  |   +   + - |
  |           |
  |   + - +   |
  | S |       |
  ------------- 
  end
\end{minted}
       \caption{Navigation}
       \label{code:navigation}
    \end{subfigure}
    \hfill
    \begin{subfigure}[b]{0.22\textwidth}
       \centering
\begin{minted}[numbersep=-4pt, escapeinside=@@]{python}
  key_and_lock: False
  icy_floor: False
  occlusion: True
  labyrinth:
  -------------  
  |   |     E |
  |   +   + - |
  |           |
  |   + - +   |
  | S |       |
  ------------- 
  end
\end{minted}
       \caption{Partial Observability}
       \label{code:partial_observability}
    \end{subfigure}
    \hfill
    \begin{subfigure}[b]{0.22\textwidth}
       \centering
\begin{minted}[numbersep=-4pt, escapeinside=@@]{python}
  key_and_lock: True
  icy_floor: False
  occlusion: False
  labyrinth:
  -------------  
  |   | D   E |
  |   +   + - |
  |           |
  |   + - +   |
  | S | K     |
  ------------- 
  end
\end{minted}
       \caption{Key and Door}
       \label{code:key_and_door}
    \end{subfigure}
    \hfill
    \begin{subfigure}[b]{0.22\textwidth}
       \centering
\begin{minted}[numbersep=-4pt, escapeinside=@@]{python}
  key_and_lock: False
  icy_floor: True
  occlusion: False
  labyrinth:
  -------------  
  |   |     E |
  |   +   + - |
  |         I |
  |   + - +   |
  | S   I   I |
  ------------- 
  end
\end{minted}
       \caption{Ice Floor}
       \label{code:ice_floor}
    \end{subfigure}
    \caption{Save files for all possible settings.}
    \label{code:all_settings}
\end{figure}

This language supports saving and loading all possible configurations (as displayed in Fig.~\ref{code:all_settings}).
Lines~\ref{code:attributes_start}-\ref{code:attributes_end}, define the setting for the labyrinth and are mutually exclusive.
The Labyrinth also only loads tiles relevant to the setting.
In other words, if all parameters are `False' but there are ice tiles (such as in Fig.~\ref{code:ice_floor} in Lines~\ref{code:structure_start}-\ref{code:structure_end}), it will not load any ice tiles.
Yet, for convenience, if the labyrinth is instantiated in one setting, Fig.~\ref{code:key_and_door} for example, and afterwards, a file with a different configuration is loaded, such as Fig.~\ref{code:ice_floor}, the labyrinth will change to load the file in its entirety.
In other words, it will overwrite the current setting (`key and door') and load the new setting (`ice floors') with the new tiles.

\noindent
Lines~\ref{code:structure_start}-\ref{code:structure_end} define the structure for each labyrinth.
They define vertical walls by using pipes `|', horizontal ones by hyphens `-', the start position $s_0$ as `S', the goal $g$ as `E', and the point between all four tiles by a plus sign `+'.
Researchers can define any navigation labyrinth by using these symbols.
It is important to note that the configuration language requires the outer walls of the labyrinth to know the size (height and width) of the labyrinth trying to be loaded.
Additionally, if the user wants to define the key and door positions ($g_k$ and $g_d$, respectively) they can do so by using `K' for key and `D' for door.
For loading ice in a specific tile, researchers can use `I' for it.
We note that if researchers define a configuration that has one of these settings turned on but does not define any location for the special tiles, the Labyrinth will define them using its internal heuristics (cf. Sec.~\ref{sec:env_types} of the main work).
Finally, the configuration supports multiple- and single-line comments by using `"""'.
So, if researchers want to comment something in the file, they can do so by opening and closing their comment with three quotation marks.

\section{Experimental Information} \label{sec:hyperparameters}

In this work we run all experiments using a machine with A$100$ NVidia GPU and $32$GB of memory.
However, the networks are relatively small, and these settings are not required to reproduce any of the results available here or in the main work.
All methods are provided by IL-Datasets~\cite{gavenski2024ildatasets}, which uses PyTorch and Adam's optimiser to update the weights.
As for learning rates, we use the default value $5 \times 10^{-4}$ for all methods.

\subsection{Network Topology}
In the main work, we experiment with six different imitation learning methods (see Sec.~\ref{sec:experiments} of the main work).
For the experiments in Tab.~\ref{tab:benchmark} of the main work, we use the original encoder from Deep Q-Network~\cite{aytar2018playing} (Tab.~\ref{tab:networks}a) and later with a ResNet-$18$~\cite{HeEtAl2016} (Tab.~\ref{tab:networks}b).
These structures are displayed in Tab.~\ref{tab:networks}.
We note that for the ResNet neural network, we follow PyTorch's implementation available at: \url{https://pytorch.org/hub/pytorch_vision_resnet/}, and the classifier layers remain the same for both encoders.

\begin{table}[H]
    \scriptsize
    \centering
    \caption{Neural network topology for encoders.}
    \label{tab:networks}
    \begin{tabular*}{\textwidth}{p{7pt} @{\extracolsep{\fill}} lcr lcr}
        \toprule
        \multirow{2}{*}[-2pt]{N} & \multicolumn{3}{c}{(a) CNN} & \multicolumn{3}{c}{(b) Resnet-18} \\ \cmidrule{2-4} \cmidrule{5-7}
            & Layer Name & Input $\times$ Output & Extra Info. & Layer Name & Input $\times$ Output & Extra Info. \\ \cmidrule{2-4} \cmidrule{5-7}
         1  & Conv2d & $3 \times 32$ & \makecell{Kernel: 8 \\Stride: 4} & Conv2d & $3 \times 64$ & \makecell{Size: 7 \\Stride:2 \\Padding: 3} \\ \cmidrule{2-4} \cmidrule{5-7}
         2  & LeakyReLU & - & Slope: 0.01 & BatchNorm2d & $64 \times 64$ & \makecell{Mom.: 0.1\\Affine: True} \\ \cmidrule{2-4} \cmidrule{5-7}
         3  & Conv2d & $32 \times 64$ & \makecell{Kernel: 4 \\ Stride: 2} & ReLU & - & - \\ \cmidrule{2-4} \cmidrule{5-7}
         4  & BatchNorm2d & $64 \times 64$ & \makecell{Mom.: 0.1 \\ Affine: True} & Max Pool & - & \makecell{Size: 3 \\Stride: 1\\Padding: 1} \\ \cmidrule{2-4} \cmidrule{5-7}
         5  & LeakyReLU & - & Slope: 0.01 & \multicolumn{3}{c}{ResNet Block ($4\times$ -- Start)}\\ \cmidrule{2-4} \cmidrule{5-7}
         6  & Conv2d & $64 \times 64$ & \makecell{Kernel: 3 \\ Stride: 1} & Conv2d & $64 \times 64$ & \makecell{Size: 3\\Stride: 1\\Padding: 1} \\ \cmidrule{2-4} \cmidrule{5-7}
         7  & BatchNorm2d & $64 \times 64$ & \makecell{Mom.: 0.1 \\Affine: True} & BatchNoorm2d & $64 \times 64$ & \makecell{Mom.: 0.1\\Affine: True} \\ \cmidrule{2-4} \cmidrule{5-7}
         8  & LeakyReLU & - & Slope: 0.01 & ReLU & - & - \\ \cmidrule{2-4} \cmidrule{5-7}
         9  & Fully Connected & $1024 \times 512$ & - & Conv2d & $64 \times 64$ & \makecell{Size: 3\\Stride: 1\\Padding: 1} \\ \cmidrule{2-4} \cmidrule{5-7}
        10  & LeakyReLU & - & Slope: 0.01 & BatchNorm2d & $64 \times 64$ & - \\ \cmidrule{2-4} \cmidrule{5-7}
        11  & Dropout & - & Prob.: $30\%$ & \multicolumn{3}{c}{ResNet Block ($4\times$ -- End)} \\ \cmidrule{2-4} \cmidrule{5-7}
        12  & Fully Connected & $512 \times 512$ & - & Fully Connected & $512 \times 512$ & - \\ \cmidrule{2-4} \cmidrule{5-7}
        13  & LeakyReLU & - & Slope: 0.01 & LeakyReLU & - & Slope: 0.01 \\ \cmidrule{2-4} \cmidrule{5-7}
        14  & Dropout & - & Prob.: $30\%$ & Dropout & - & Prob.: $30\%$ \\ \cmidrule{2-4} \cmidrule{5-7}
        15  & Fully Connected & $512 \times 4$ & - & Fully Connected & $512 \times 512$ & - \\ \cmidrule{2-4} \cmidrule{5-7}
        16 & & & & LeakyReLU & - & Slope: 0.01 \\ \cmidrule{5-7}
        17 & & & & Dropout & - & Prob.: $30\%$ \\ \cmidrule{5-7}
        18 & & & & Fully Connected & $512 \times 4$ & - \\
        \bottomrule       
    \end{tabular*}
\end{table}

Finally, for GAIL's discriminator, we use a simple visual encoder (displayed in Tab~\ref{tab:gail}).

\begin{table}[H]
    \scriptsize
    \centering
    \caption{Discriminator Topology}
    \label{tab:gail}
    \begin{tabular}{c lcr}
        \toprule
        \multirow{2}{*}[-2pt]{N} & \multicolumn{3}{c}{GAIL's Discriminator} \\ \cmidrule{2-4}
                           & Layer Name & Input $\times$ Output & Extra Info. \\ \cmidrule{2-4}
        1 & Conv2d & $3 \times 32$ & \makecell{Size: 8 and Stride: 4} \\ \cmidrule{2-4}
        2 & LeakyReLU & - & Slope: 0.01 \\ \cmidrule{2-4}
        3 & Conv2d & $32 \times 64$ & \makecell{Size: 4 and Stride: 2} \\ \cmidrule{2-4}
        4 & LeakyReLU & - & Slope: 0.01 \\ \cmidrule{2-4}
        5 & Conv2d & $64 \times 64$ & \makecell{Size: 3 and Stride: 1} \\ \cmidrule{2-4}
        6 & LeakyReLU & - & Slope: 0.01 \\ \cmidrule{2-4}
        7 & Fully Connected & $1028 \times 256$ & - \\ \cmidrule{2-4}
        8 & LeakyReLU & - & Slope: 0.01 \\ \cmidrule{2-4}
        9 & Fully Connected & $256 \times 256$ & - \\ \cmidrule{2-4}
        10 & LeakyReLU & - & Slope: 0.01 \\ \cmidrule{2-4}
        11 & Fully Connected & $256 \times 1$ & - \\
        \bottomrule
    \end{tabular}
\end{table}

\section{Extra Benchmark Results}

This section provides a detailed breakdown of agent performance across increasingly complex Labyrinth sizes.
By isolating each configuration, we assess how well imitation learning methods generalise beyond the training distribution as structural diversity and trajectory length increase.

\subsection{Results for $\mathbf{3 \times 3}$ Labyrinths}

We expect labyrinths that are $3 \times 3$ of size to be an easier challenge to imitation learning methods.
As stated in the main work in Sec.~\ref{sec:structure}, labyrinths of smaller size have higher chances of presenting similar trajectories to their goal.
Therefore, imitation learning methods, even with smaller encoders, are expected to achieve higher $SR$.

\begin{table}[H]
    \tiny
    \centering
    \caption{Results for training, validation and testing splits for the navigation task in a $3 \times 3$ Labyrinth.}
    \label{tab:benchmark3x3}
    \begin{tabular*}{\textwidth}{@{\extracolsep{\fill}} lc r r r r r r}
        \toprule
        Splits & Metric
            & \multicolumn{1}{c}{BC}  
            & \multicolumn{1}{c}{DAgger}
            & \multicolumn{1}{c}{GAIL}  
            & \multicolumn{1}{c}{BCO}  
            & \multicolumn{1}{c}{SQIL}  
            & \multicolumn{1}{c}{IUPE} \\ \midrule
        \multirow{2}{*}{Train} & $AER$ & $-1.36 \pm 2.48$ & $-0.74 \pm 2.35$ & $-1.03 \pm 2.45$ & $\mathbf{-0.59 \pm 2.30}$ & $-2.50 \pm 2.27$ & $-1.22 \pm 2.45$ \\
                               & $SR$ &$53.12\%$ & $65.62\%$ & $59.38\%$ & $\mathbf{68.75\%}$ & $31.25\%$ & $56.25\%$ \\
        \multirow{2}{*}{Valid.} & $AER$ & $\mathbf{-0.59 \pm 2.30}$ & $-1.06 \pm 2.43$ & $-1.50 \pm 2.5$ & $-1.36 \pm 2.48$ &  $-3.24 \pm 1.77$ & $-0.91 \pm 2.39$ \\
                                & $SR$ & $\mathbf{68.75\%}$ & $59.38\%$ & $50.00\%$ & $53.12\%$ & $15.62\%$ & $62.50\%$ \\
        \multirow{2}{*}{Test} & $AER$ & $\mathbf{-0.74 \pm 2.36}$ & $-1.20 \pm 2.45$ & $-1.22 \pm 2.45$ & $-0.90 \pm 2.40$ & $-2.93 \pm 2.02$ & $-0.92 \pm 2.39$ \\
                              & $SR$ & $\mathbf{65.62\%}$ & $56.25\%$ & $56.25\%$ & $62.50\%$ & $21.87\%$ & $62.50\%$ \\
        \multirow{2}{*}{$s_0$} & $AER$ & $-1.20 \pm 1.65$ & $-1.36 \pm 2.48$ & $-0.87 \pm 2.42$ & $-1.04 \pm 2.44$ & $-2.31 \pm 2.27$ & $\mathbf{-0.45 \pm 2.23}$ \\
                               & $SR$ & $56.25\%$ & $53.12\%$ & $62.50\%$ & $59.38\%$ & $34.37\%$ & $\mathbf{71.87\%}$ \\
        \multirow{2}{*}{$s_0$ and $g$} & $AER$ & $-3.38 \pm 1.64$ & $-3.69 \pm 1.20$ & $-3.37 \pm 1.65$ & $-3.22 \pm 1.80$ & $\mathbf{-2.47 \pm 2.27}$ & $-3.69 \pm 1.20$ \\
                                       & $SR$ & $12.50\%$ & $6.25\%$ & $12.50\%$ & $15.62\%$ & $\mathbf{31.25\%}$ & $6.25\%$ \\
        \bottomrule
    \end{tabular*}
\end{table}

Table~\ref{tab:benchmark3x3} shows that a smaller labyrinth size indeed facilitates imitation learning.
The pure imitation learning supervised approaches achieve the highest $SR$.
SQIL consistently underperforms across splits, with the exception of unseen conditions ($s_0$ and $g$), where it achieves $31.25\% $ $SR$.
IUPE demonstrates the highest success rate on $s_0$, suggesting better robustness in generalisation to novel initial states, but drastically underperforms on different initial states and goals.
We hypothesise that the shift over the actions (to a more uniform one, as stated in Sec.~\ref{sec:generalisation}) is too drastic for these methods to generalise and achieve the goal.
These results highlight that while simple environments boost overall performance, generalisation remains a core challenge, with only a subset of methods, notably IUPE and BCO (self-supervised imitation learning methods), showing promise beyond the training distribution.

\subsection{Results for $\mathbf{4 \times 4}$ Labyrinths}

As we scale the environment to a $4\times 4$ labyrinth, the imitation learning task becomes noticeably more challenging. 
Unlike the $3\times 3$ case, where the overlap of many trajectory aids generalisation, larger labyrinths induce more structural and strategic diversity, reducing the benefit of purely behavioural memorisation.

\begin{table}[H]
    \tiny
    \centering
    \caption{Results for training, validation and testing splits for the navigation task in a $4 \times 4$ Labyrinth.}
    \label{tab:benchmark4x4}
    \begin{tabular*}{\textwidth}{@{\extracolsep{\fill}} lc r r r r r r}
        \toprule
        Splits & Metric
            & \multicolumn{1}{c}{BC}  
            & \multicolumn{1}{c}{DAgger}
            & \multicolumn{1}{c}{GAIL}  
            & \multicolumn{1}{c}{BCO}  
            & \multicolumn{1}{c}{SQIL}  
            & \multicolumn{1}{c}{IUPE} \\ \midrule
        \multirow{2}{*}{Train} & $AER$ & $-2.29 \pm 2.36$ & $-2.29 \pm 2.36$ & $-2.35 \pm 2.35$ & $-2.29 \pm 2.36$ & $-2.32 \pm 2.32$ & $\mathbf{-1.53 \pm 2.47}$ \\
                               & $SR$ & $34.00\%$ & $34.00\%$ & $33.00\%$ & $34.00\%$ & $34.00\%$ & $\mathbf{50.00\%}$ \\
    \multirow{2}{*}{Valid.} & $AER$ & $-2.29 \pm 2.36$ & $-2.14 \pm 2.4$ & $-2.29 \pm 2.36$ & $-2.29 \pm 2.36$ & $-3.24 \pm 1.77$ & $\mathbf{-0.76 \pm 2.34}$ \\
                                & $SR$ & $34.00\%$ & $37.00\%$ & $34.00\%$ & $34.00\%$ & $15.00\%$ & $\mathbf{65.00\%}$ \\
    \multirow{2}{*}{Test} & $AER$ & $-2.14 \pm 2.4$ & $-2.29 \pm 2.36$ & $-2.20 \pm 2.40$ & $-2.14 \pm 2.4$ & $-3.38 \pm 1.63$ & $\mathbf{-0.61 \pm 2.29}$ \\
                              & $SR$ & $37.00\%$ & $34.00\%$ & $36.00\%$ & $37.00\%$ & $12.00\%$ & $\mathbf{68.00\%}$ \\
        \multirow{2}{*}{$s_0$} & $AER$ & $-1.82 \pm 2.47$ & $-1.67 \pm 2.48$ & $-1.80 \pm 0.99$ & $\mathbf{-1.51 \pm 2.49}$ & $-2.61 \pm 2.22$ & $-1.53 \pm 2.47$ \\
                               & $SR$ & $43.00\%$ & $46.00\%$ & $44.00\%$ & $\mathbf{50.00\%}$ & $28.00\%$ & $50.00\%$ \\
        \multirow{2}{*}{$s_0$ and $g$} & $AER$ & $-3.69 \pm 1.2$ & $-3.84 \pm 0.87$ & $-3.80 \pm 0.39$ & $-3.84 \pm 0.87$ & $\mathbf{-2.92 \pm 2.04}$ & $-3.07 \pm 1.93$ \\
                                       & $SR$ & $6.00\%$ & $3.00\%$ & $4.00\%$ & $3.00\%$ & $\mathbf{21.00\%}$ & $18.00\%$ \\
        \bottomrule
    \end{tabular*}
\end{table}

Table~\ref{tab:benchmark4x4} illustrates this transition, with most methods plateauing at relatively low success rates across all splits.
Traditional behavioural approaches -- BC, DAgger, and GAIL -- achieve comparable performance during training ($SR \approx 34\%$) and show minimal gains on validation and test sets, indicating a limited capacity to generalise beyond memorised demonstrations.
In contrast, IUPE substantially outperforms all baselines in validation ($65\%$) and test ($68\%$), with consistent superiority across every evaluation split.
This suggests that the inductive biases introduced by IUPE — likely due to its inverse dynamics structure and self-supervised training scheme yields a more robust understanding of the environment's transition dynamics, enabling better policy adaptation to unseen instances. 
Notably, even in the more difficult $s_0$ and $g$ setting, IUPE achieves a $50\%$ $SR$, outperforming all methods except BCO.
Meanwhile, SQIL continues to struggle, especially in generalisation scenarios:
it achieves the lowest SR on validation ($15\%$) and test ($12\%$), although it obtains the highest SR ($21\%$) in the most challenging $s_0$ and $g$ setting.
This anomalous result may be due to the reward-shaping mechanism in SQIL biasing action selection towards diverse outcomes, which occasionally aligns with goal-directed behaviour under novel conditions (especially in the case of smaller labyrinths).
Overall, Table~\ref{tab:benchmark4x4} highlights the fragility of standard imitation learning methods under modestly increased environment complexity and reinforces the promise of self-supervised and model-based techniques for achieving broader generalisation.

\subsection{Results for $\mathbf{5 \times 5}$ Labyrinths}

The $5\times 5$ environment represents the most complex setting in our benchmark, with longer optimal trajectories, sparser rewards, and greater variation in structure.

\begin{table}[th]
    \tiny
    \centering
    \caption{Results for training, validation and testing splits for the navigation task in a $5 \times 5$ Labyrinth.}
    \label{tab:benchmark5x5}
    \begin{tabular*}{\textwidth}{@{\extracolsep{\fill}} lc r r r r r r}
        \toprule
        Splits & Metric
            & \multicolumn{1}{c}{BC}  
            & \multicolumn{1}{c}{DAgger}
            & \multicolumn{1}{c}{GAIL}  
            & \multicolumn{1}{c}{BCO}  
            & \multicolumn{1}{c}{SQIL}  
            & \multicolumn{1}{c}{IUPE} \\ \midrule
        \multirow{2}{*}{Train} & $AER$ & $-2.11 \pm 2.41$ & $-1.18 \pm 2.45$ & $-0.98 \pm 1.89$ & $-0.53 \pm 2.23$ & $-3.80 \pm 0.96$ & $\mathbf{0.27 \pm 2.39}$ \\
            & $SR$ & $37\%$ & $57\%$ & $61\%$ & $70\%$ & $4\%$ & $\mathbf{75\%}$ \\
        \multirow{2}{*}{Valid.} & $AER$ & $-3.70 \pm 1.18$ & $-3.75 \pm 1.08$ & $-3.57 \pm 1.58$ & $-3.90 \pm 0.69$ & $-3.95 \pm 0.49$ & $\mathbf{-2.80 \pm 2.12}$ \\
            & $SR$ & $6\%$ & $5\%$ & $9\%$ & $2\%$ & $1\%$ & $\mathbf{21\%}$ \\
        \multirow{2}{*}{Test} & $AER$ & $-3.90 \pm 0.70$ & $-3.80 \pm 0.97$ & $-3.85 \pm 0.85$ & $-3.85 \pm 0.85$ & $-4.00 \pm 0.00$ & $\mathbf{-3.85 \pm 1.00}$ \\
            & $SR$ & $2\%$ &$4\%$ & $3\%$ & $3\%$ & $0\%$ & $\mathbf{5\%}$ \\
        \multirow{2}{*}{$s_0$} & $AER$ & $-2.00 \pm 0.98$ & $-0.90 \pm 0.97$ & $-0.60 \pm 0.92$ & $-0.35 \pm 0.88$ & $-3.55 \pm 0.57$ & $\mathbf{0.00 \pm 0.80}$ \\
                               & $SR$ & $40.00\%$ & $62.00\%$ & $68.00\%$ & $73.00\%$ & $9.00\%$ & $\mathbf{80.00\%}$ \\
        \multirow{2}{*}{$s_0$ and $g$} & $AER$ & $-4.00 \pm 0.00$ & $-4.00 \pm 0.00$ & $-4.00 \pm 0.00$& $-4.00 \pm 0.00$ & $-4.00 \pm 0.00$ & $-4.00 \pm 0.00$ \\
                                                & $SR$ & $0.00\%$ & $0.00\%$ & $0.00\%$ & $0.00\%$ & $0.00\%$ & $0.00\%$ \\
        \bottomrule
    \end{tabular*}
\end{table}

As shown in Table~\ref{tab:benchmark5x5}, this setting induces significant degradation in performance across all methods, especially in generalisation.
Training success rates remain relatively high for methods such as IUPE ($75\%$) and BCO ($70\%$), indicating that their self-supervised approach is capable of fitting the training distribution well.
However, this performance does not transfer to unseen configurations: validation success rates drop to $21\%$ and $2\%$ respectively, and test results fall even further, with IUPE reaching only $5\%$ and BCO stagnating at $3\%$.
This steep decline underscores the increasing gap between memorisation and generalisation as environment complexity grows.

Most striking is the complete failure of all the methods in the $s_0$ and $g$ split, where the success rates uniformly drop to $0\%$.
This clearly illustrates that none of the evaluated methods, including those with stronger inductive structures such as IUPE or BCO, can act meaningfully when both the initial and goal states are novel.
Although IUPE achieves the highest SR in the $s_0$ setting ($80\%$), indicating a capacity to generalise from unseen starting positions alone, its inability to deal with novel goals further confirms the limitations of current methods in reasoning about goal-conditioned policies.
The failure of SQIL is especially significant, performing poorly across all splits, achieving just $4\%$ in training and $\approx 0\%$ elsewhere, including both generalisation conditions.
This suggests that its reliance on implicit Q-learning from suboptimal reward signals is particularly ill-suited for environments with high state aliasing and long planning horizons.

\subsection{Extra Results Conclusion}
Across the three levels of environment complexity, a consistent trend emerges.
Although most methods perform reasonably well under training conditions and modest distribution shifts, their generalisation rapidly degrades as task structure becomes more diverse.
In the simplest $3\times 3$ setting, even behavioural approaches perform competitively; but by the time we reach the $5\times 5$ environments, success is effectively unattainable for all methods on the most challenging splits.

The strongest generalisation comes from IUPE, followed by BCO.
Both of which leverage self-supervised objectives or explicit modelling of transitions.
These methods show meaningful robustness across multiple splits, especially in intermediate conditions like unseen initial states.
Still, none of the evaluated techniques succeed when initial and goal configurations change, highlighting the lack of robustness of current imitation learning strategies when exposed to true task-level variability.

These findings underscore the pressing need for novel approaches in imitation learning that go beyond local behaviour matching.
Promising directions include goal-conditioned modelling, compositional planning, or hybrid methods that blend imitation with exploratory or model-based learning components.
Ultimately, our benchmark reveals that while simple environments can be mastered through memorisation and interpolation, scalable generalisation demands a fundamentally different approach.

\section{Extra Results for Most Used Environments}

Table~\ref{tab:seeds} in the main work provides a summary of the first-state distribution across $100,000$ initialisations for common imitation learning environments, using Manhattan distance as a proxy for distributional similarity.
These results show that environments like MountainCar, CartPole, and Hopper exhibit extremely low variation in initial states, suggesting that agents are often evaluated under conditions very similar to those seen during training.
To complement this, we conducted additional experiments to measure distributional shifts in both action and trajectory (state) distribution between training and evaluation.
For discrete environments such as Labyrinth, CartPole, and MountainCar, we used the Jensen-Shannon (JS) distance, which is well-suited for comparing categorical distributions like discrete action spaces.
For continuous control environments (e.g., Hopper, Walker, and HalfCheetah), we used the Wasserstein (WS) distance for both action and trajectory distributions, as it better captures geometric differences in continuous spaces. We note that we also use the WS distance for the Labyrinth state space to make the comparison between values fairer.

The results from Table~\ref{tab:metrics_heatmap} work as a proxy for the heatmap images from Figure~\ref{fig:lab-divergence} from our main work.
Moreover, we display the average reward to show that the imitation learning agents are not performing poorly, but rather that they are achieving a reward close to their teachers, and that the low difference is not a result of them failing early in an episode.
The table below shows that Labyrinth exhibits significantly higher distributional divergence, with an action distribution difference of $0.0165$ and a trajectory difference of $0.0475$.
Compared to the other benchmarks, where values are often near zero, these results reinforce our claim that Labyrinth presents a more rigorous generalisation challenge, even in small maze configurations.
While this analysis emphasises statistical significance via p-values, it is important to contextualise these values alongside the magnitude of behavioural change.
For example, environments such as CartPole and MountainCar show very low p-values despite minimal trajectory differences ($0.0035$ and $0.0037$, respectively), indicating that even minor deviations are statistically consistent but not necessarily impactful.
In contrast, Hopper exhibits both a low trajectory difference ($0.0023$) and a non-significant p-value ($0.924$), suggesting stable behaviour.
Labyrinth, however, combines a large trajectory difference with a highly significant p-value, providing strong evidence of meaningful behavioural divergence.

\begin{table}[h]
    \scriptsize
    \centering
    \caption{Action and trajectory difference over $1 \times 10^5$ seeds.}
    \label{tab:metrics_heatmap}
    \begin{tabular*}{\textwidth}{l@{\extracolsep{\fill}}rrrr}
        \toprule
        Environment & Average Reward & Action Distribution Difference & Average Trajectory Difference & Trajectory p-value  \\
        \midrule
        Labyrinth   &    0.9535 & 0.0165 & 0.0475 & 0.000 \\
        MountainCar & -101.1170 & 0.0093 & 0.0035 & 0.011 \\
        CartPole    &  500.0000 & 0.0000 & 0.0037 & 0.000 \\
        Hopper      & 3530.6376 & 0.0007 & 0.0023 & 0.924 \\
        Walker      & 4713.9521 & 0.0005 & 0.0016 & 1.000 \\
        HalfCheetah & 9526.0863 & 0.0014 & 0.0103 & 0.254 \\
        \bottomrule
    \end{tabular*}
\end{table}

\section{On the Difference from Procedurally Generated Datasets}

Machine learning and reinforcement learning benefit from standardised benchmarking datasets and environments, which enable consistent comparisons across methods and ensure reproducibility. 
These benchmarks often include hidden test sets to preserve evaluation integrity.
In contrast, imitation learning lacks such standardisation.
Imitation learning studies frequently rely on procedurally generated datasets, which cannot be fully reproduced, making it impossible to compare results fairly.
Researchers often need to generate new teacher datasets, locate or reimplement baseline methods, and tune hyperparameters, all of which introduce variability and potential bias.
This lack of uniform benchmarks complicates evaluation and reinforces the existing challenges in imitation learning testing practices.

Environments, such as MiniGrid~\citep{chevalier2023minigrid}, Procgen~\cite{cobbe2020leveraging}, and Crafter~\cite{hafner2021benchmarking}, are valuable for evaluating generalisation in reinforcement learning and, in principle, could be used for imitation learning as well.
However, imitation learning faces unique challenges in terms of achieving generalisation, despite using the same environments.
This stems from the fact that imitation learning agents are trained solely on expert demonstrations and do not interact with the environment during training (in offline settings).
As a result, they lack exposure to out-of-distribution states and must rely entirely on the coverage and diversity of the training trajectories.
In contrast, reinforcement learning agents are free to explore and adapt during training, allowing them to recover from unfamiliar states and improve upon suboptimal demonstrations.
The difference in training dynamics means that generalisation failures in imitation learning are often more severe and harder to diagnose, especially when the evaluation protocol lacks control over the distributional shift between training and test environments (one of the key differences between Labyrinth and the other environments not specifically designed for imitation learning, such as the ones you mentioned).

Regarding the use of hidden test sets: while reinforcement learning benchmarks like Procgen and Crafter do use held-out seeds for evaluation, they typically sample environments randomly, which can lead to inconsistent generalisation pressure across runs.
Labyrinth addresses this by offering explicit control over structural variation, start and goal positions, and task complexity (allowing users to isolate specific generalisation factors and reproduce experiments precisely).
Another advantage of Labyrinth over non-imitation learning environments.

To your question about whether there are benchmarks solvable by reinforcement learning but not imitation learning: an illustrative example is the MetaWorld benchmark~\cite{metaworld}.
MetaWorld comprises a suite of robotic manipulation tasks with procedural variations in object positions and goals. 
Reinforcement learning agents trained with exploration and reward feedback, such as those using meta-learning or model-based strategies, have demonstrated strong generalisation across unseen tasks (approximately $60\%$ for meta-reinforcement learning according to~\citeauthor{metaworld}).
In contrast, imitation learning agents trained solely on expert demonstrations often fail to generalise in MetaWorld~\cite{difftori,isimitation}, especially when the demonstrations do not fully cover the diversity of initial states or task configurations.
\citeauthor{difftori} achieved an average of $42\%$ across multiple tasks, with their baselines Behavioural Cloning achieving $34\%$ (half of the reinforcement learning results), and \citeauthor{isimitation} achieved an average around $32\%$. 

Moreover, \citeauthor{isimitation} notes:
\begin{quote}
We can infer from this that imitation learning alone may not suffice to build a truly general-purpose model, particularly when aiming to tackle tasks that span a broad range of domains. Even within a single domain, variations in embodiments, scenes, and instructions can pose significant challenges.
\end{quote}
We attribute this to the fact that imitation learning agents lack the ability to recover from unfamiliar states or adapt to novel goal during training.
For us, MetaWorld exemplifies a benchmark where reinforcement learning succeeds due to its interactive learning paradigm, while imitation learning struggles without sufficient coverage or structural reasoning.
This contrast reinforces the motivation behind Labyrinth: to provide a controlled and interpretable environment where generalisation failures in imitation learning can be studied in isolation from exploration and reward shaping.

Finally, Labryinth and MiniGrid are both grid-based and support procedural generation. We have already pointed out some differences between them, but we can briefly enumerate Labyrinth's unique features as follows:
\begin{enumerate}
    \item \textit{Full observability and known optimal actions}, enabling fine-grained evaluation and debugging;
    \item \textit{Explicit separation of training, validation, and test sets}, with guarantees of structural uniqueness;
    \item \textit{Control over generalisation complexity and distribution shifts}, allowing for better evaluation of imitation learning methods generalisation capabilities; and
    \item \textit{A reproducible configuration language and dataset generation tools}, tailored for imitation learning benchmarking.
\end{enumerate}
We believe these features make Labyrinth uniquely suited for studying generalisation in imitation learning, while also providing a foundation for reinforcement learning evaluation in future work.

\end{document}